%% file: main.tex
\documentclass{article} 
\usepackage{arxiv,times}

\input{math_commands.tex}

\usepackage{hyperref}
\usepackage{url}
\usepackage{lipsum}
\usepackage[utf8]{inputenc}
\usepackage[T1]{fontenc}
\usepackage{hyperref}
\usepackage{url}            
\usepackage{booktabs}       
\usepackage{amsfonts}       
\usepackage{nicefrac}       
\usepackage{algorithm,algpseudocode}
\usepackage{amsfonts}
\usepackage{multicol}
\usepackage{amsmath}
\usepackage{amsthm}
\usepackage{amssymb}
\usepackage{tikz}\usetikzlibrary{arrows.meta,positioning,fit,calc,backgrounds}
\usepackage[title]{appendix}
\usepackage{bm}
\usepackage{courier}
\usepackage{enumitem}
\usepackage{graphicx}
\usepackage{float}
\usepackage{url}
\usepackage{multirow}
\usepackage[dvipsnames]{xcolor}
\usepackage{epsfig}
\usepackage{graphicx}
\usepackage{subcaption}
\usepackage{wrapfig}   
\usepackage{caption}   
\usepackage{microtype}
\usepackage{tikz}
     \definecolor{redteam}{RGB}{178,40,44}
     \definecolor{blueteam}{RGB}{38,84,152}
     \definecolor{boxgray}{RGB}{242,242,242}
     \definecolor{boxgold}{RGB}{252,249,240}
\usepackage[most]{tcolorbox}
\usepackage{listings}
\usepackage{lstlinebgrd}
\definecolor{cmtgreen}{RGB}{0,128,0}
\definecolor{kwblue}{RGB}{0,0,180}
\definecolor{strbrown}{RGB}{163,21,21}
\definecolor{hlbg}{RGB}{226,226,244}
\newcommand{\hlr}[2]{\ifnum\value{lstnumber}>\numexpr#1-1\relax
  \ifnum\value{lstnumber}<\numexpr#2+1\relax\color{hlbg}\fi\fi}
\lstdefinestyle{dcapy}{
  language=Python, basicstyle=\ttfamily\tiny, numbers=left,
  numberstyle=\ttfamily\tiny\color{black!55}, numbersep=5pt,
  keywordstyle=\bfseries\color{kwblue},
  commentstyle=\itshape\color{cmtgreen},
  stringstyle=\color{strbrown},
  showstringspaces=false, breaklines=true, breakautoindent=true,
  prebreak=\mbox{\tiny$\hookrightarrow$}, breakindent=8pt,
  xleftmargin=13pt, aboveskip=2pt, belowskip=2pt, frame=tb, framesep=3pt,
}

\newcommand{\redagent}{Discriminator}

\newcommand{\blueagent}{Generator}

\newif\ifgalpack
\newcommand{\galpanelht}{4.35cm}
\newcommand{\galimght}{3.55cm}
\newcommand{\galbarht}{1.5cm}

\title{ Generative Adversarial Loops}

\author{%
Kislay Aditya Oj\textsuperscript{$\alpha$,*} \quad
Nidhi Jain\textsuperscript{$\beta$,*} \quad
Sri Surya Varma Datla\textsuperscript{$\alpha$} \quad
Priyanka Jayaswal\textsuperscript{$\beta$} \\
\textbf{Kumar Krishna Agrawal\textsuperscript{$\gamma$}} \quad
\textbf{Aditya Desai\textsuperscript{$\alpha$}}
}

\iclrfinalcopy 
\begin{document}

\maketitle
\begingroup
\makeatletter
\long\def\@makefntext#1{\noindent#1}
\makeatother
\renewcommand{\thefootnote}{\fnsymbol{footnote}}
\footnotetext{%
\textsuperscript{$\alpha$} Computer Science and Engineering, Indian Institute of Technology Bombay, Mumbai, India \\
\textsuperscript{$\beta$} Independent researcher \\
\textsuperscript{$\gamma$} Electrical Engineering and Computer Sciences, University of California, Berkeley, USA \\
\textsuperscript{*} Equal contribution.
}
\endgroup

\begin{abstract}
\input{sections/abstract_3_archive}
\end{abstract}
\vspace{-0.2em}
\section{Introduction} \label{sec:introduction}
\vspace{-0.2em}
\input{sections/introduction_2}

\vspace{-0.2em}
\section{Related Work} \label{sec:related}
\vspace{-0.2em}
\input{sections/related-work}
\vspace{-0.2em}
\section{Generative Adversarial Loops}
\vspace{-0.2em}
\input{sections/method_1}
\vspace{-0.2em}
\section{Experiments on Efficiency Problems in AI Inference}
\vspace{-0.2em} 
\input{sections/case_studies_compact}

\vspace{-0.2em}
\section{Conclusion}
\vspace{-0.2em}
In this work, we present Generative Adversarial Loops (GALs), a framework for automated algorithm discovery through the interaction of generation and discrimination. While prior work has largely automated the generative component, we argue that the discriminator is equally important and underexplored. We instantiate GALs using GEPA, Cursor, and Claude Agents on approximation problems in efficient AI inference. GALs uncover interpretable failure patterns in existing algorithms and, when coupled with the generative loop, use these failures to discover improved algorithms that generalize across diverse benchmarks. More broadly, GALs can become a useful tool for researchers, accelerating algorithmic iteration while providing a cost certificate for robustness: the adversarial effort or computational cost required to expose a failure mode. While this certificate does not establish universal robustness, it quantitatively measures how difficult a method is to break under a specified adversarial generation and evaluation procedure.





\subsubsection*{Acknowledgments}
This work was primarily supported by the IIT Bombay Seed Grant (RD/0526-IRCCSH0-003). We thank BharatGen for providing access to H200 GPU compute resources that supported the experiments in this work. We are grateful for their support and resources.

\bibliography{iclr2027_conference}
\bibliographystyle{iclr2027_conference}
\clearpage
\appendix
\section{Appendix}
\input{sections/appendix}

\end{document}

%% file: math_commands.tex
\usepackage{amsmath,amsfonts,bm}

\def\eqref#1{equation~\ref{#1}}

\def\1{\bm{1}}

\DeclareMathAlphabet{\mathsfit}{\encodingdefault}{\sfdefault}{m}{sl}
\SetMathAlphabet{\mathsfit}{bold}{\encodingdefault}{\sfdefault}{bx}{n}



%% file: sections/abstract_3_archive.tex
AI research progress can be viewed as the interaction between two processes: benchmark creation and method discovery. Historically, both were driven by human intelligence. However, recent advances in AI have accelerated automated method discovery, while automated benchmark creation has received comparatively less attention. To enable self-advancing systems, we propose Generative Adversarial Loop (GAL), a generator-discriminator framework alternating between two agentic searches: (1) a discriminator that generates adversarial data to expose weaknesses in current systems, and (2) a generator that discovers algorithms to overcome them. We apply this framework to approximation algorithms for efficient inference. Unlike existing auto research systems, which primarily focus on algorithm discovery, GAL introduces a discriminator agent that automates goalpost setting by continually searching for weaknesses in the current algorithm. We demonstrate adversarial data generation across four tasks: KV compression, sparse video generation, sparse attention, and context extension, where the discriminator identifies weaknesses in state of the art techniques. We further show that GAL enables autonomous improvement, with newly discovered algorithms improving not only on adversarially generated data, but also on established benchmarks.  Specifically, GAL improves CompactorPress on KV compression with Qwen3-4B at 4x, raising performance on the discriminator dataset from 0.35 to 0.97, while also 
outperforming RULER-HARD (+0.77 pts). For context extension, GAL boosts Dual Chunk Attention from 0.20 to 0.90 on the discriminator dataset, while yielding gains on  standard benchmarks(ScienceFiction (+6 pts) and PG19 32K (-0.33 PPL)). GAL thus provides a path toward autonomous goalpost setting and algorithmic improvement, where AI systems  continually discover their own weaknesses and develop methods to overcome them.

%% file: sections/introduction_2.tex
\looseness=-1 Research progress in machine learning can broadly be understood as the interaction between two complementary processes: benchmark creation and method discovery. Benchmark creation sets measurable goals that expose limitations in current systems, while method discovery develops algorithms, architectures, and optimization techniques to overcome them. From this perspective, benchmark creation is discriminative: it identifies failure cases, whereas method discovery is generative: it produces solutions. Progress emerges through an alternating cycle between the two: new benchmarks expose capability gaps, enabling focused algorithmic innovation; as methods improve, benchmarks eventually saturate, motivating more challenging evaluations. This cycle has repeatedly shaped machine learning, from MNIST~\citep{lecun1998gradient}, CIFAR~\citep{krizhevsky2009learning}, and ImageNet~\citep{russakovsky2015imagenet} in image recognition to GLUE~\citep{wang2018glue}, SuperGLUE~\citep{wang2019superglue}, MMLU~\citep{hendrycks2020measuring}, and GSM8K~\citep{cobbe2021training} in language understanding and reasoning. Sustaining progress therefore requires continuously creating benchmarks that are more challenging, diverse, and discriminative. Yet both sides of this cycle have historically remained largely human driven: researchers design evaluations to expose new weaknesses and develop methods to address them.

\looseness=-1 Recently, this balance between benchmark creation and method discovery has begun to shift. AI systems increasingly contribute to scientific discovery through rapid prototyping, automated experimentation, large-scale evaluation, and hypothesis generation. Notable successes include AlphaTensor~\citep{fawzi2022discovering}, which discovered improved matrix multiplication algorithms, and AlphaFold~\citep{jumper2021highly}, which enabled large-scale protein structure prediction. This progress has motivated the development of AI-Driven Research Systems (ADRS)~\citep{cheng2025barbarians}. Systems such as AlphaEvolve~\citep{novikov2025alphaevolve} and OpenEvolve~\citep{openevolve} use iterative search and automated evaluation to discover algorithmic improvements. Broader evolutionary and agentic approaches, such as GEPA~\citep{agrawal2026gepa}, further expand this paradigm. However, existing ADRS approaches automate only one side of the research cycle: method discovery. Nearly all still rely on human-designed benchmarks and evaluation protocols to define progress. As method discovery becomes increasingly automated and scalable, benchmark construction becomes the bottleneck. The next frontier of automated scientific discovery is therefore not only discovering better methods, but also automatically discovering what to evaluate.

\looseness=-1 This paper addresses this emerging bottleneck by introducing Generative Adversarial Loops (GAL), a framework that automates discriminative benchmark discovery alongside generative method discovery. Unlike the ADRS paradigm, which focuses primarily on the generative loop, GAL automates both processes. Benchmark discovery, however, presents unique challenges. Unlike method discovery, where benchmarks define the objective to optimize, creating meaningful benchmarks often requires reliable ground truth and the ability to identify regions beyond the capabilities of current methods. This dependence has historically made benchmark design a human-driven process. We observe that many important research domains admit ground truth by construction. In such settings, benchmark generation can itself become algorithmic: datasets can be synthesized automatically while preserving reliable supervision, enabling systematic discovery of failure cases and controlled stress-testing of state-of-the-art systems.

\looseness=-1 One such domain is efficient inference through approximation. To address system bottlenecks in AI inference, many approximation algorithms have been proposed. This work focuses on four such problems: (1) \textbf{KV Cache Compression}~\citep{kvpress2024, liu2023scissorhands, zhang2023ho}: compressing the KV cache of autoregressive models in embedding space to address the memory bottleneck in LLM inference engines; (2) \textbf{Sparse Decoding}~\citep{tang2024quest, hooper2024fast, desai2026vattention}: sparsifying attention computation during decoding to address the memory bandwidth bottleneck of autoregressive decoding; (3) \textbf{Sparse Video Generation}~\citep{xi2025sparse, yang2026sparse}: sparsifying the prefill workload of diffusion models to enable faster video generation; and (4) \textbf{Context Extension}~\citep{an2024training, liu2025reattention}: enabling training-free context extension, allowing models to process contexts substantially longer than their training context window. This domain naturally provides a setting for automated benchmark generation for two reasons. First, benchmarks can often be generated programmatically. For example, long-context benchmarks can be constructed by systematically placing relevant pieces of information at different locations throughout a long context. Second, more importantly, the unapproximated model provides a natural source of ground truth. For example, whether a prompt is suitable for evaluating sparse video generation can be determined by the quality of the video produced by the dense, unapproximated model. These properties enable us to automatically generate benchmarks while retaining reliable supervision for evaluating approximation algorithms.

\looseness=-1 We formalize benchmark evolution as an iterative process of accumulating data samples, where each iteration discovers a new task on which the state-of-the-art approximation algorithm fails to maintain fidelity to the unapproximated baseline. To discover such failures, we leverage recent autonomous evolution algorithms such as GEPA. When a new failure pattern is identified, we use the ADRS paradigm to evolve the approximation algorithm on the expanded benchmark. Thus, benchmark discovery and method discovery form an alternating loop, with both stages leveraging the same underlying machinery of evolutionary search. More broadly, we argue that automated benchmark generation represents a complementary axis to automated method discovery, which is already gaining momentum, and offers a scalable mechanism for sustaining accelerated research progress.

The focus of this paper is to demonstrate the utility of automated benchmark generation for discovering weaknesses in state-of-the-art algorithms that have seemingly saturated existing public benchmarks and coupling it with algorithm evolution to show possibility of complete autonomous progress. To this end, we show that GALs can uncover weaknesses in four algorithms: FastKVZip~\citep{kim2026fast} for KV cache compression, PQCache~\citep{zhang2025pqcache} for sparse attention, SVG~\citep{xi2025sparse} for sparse video generation, and DCA~\citep{an2024training} for context extension. Most of the adversarial failure modes uncovered by GALs are interpretable, providing insight into the limitations of existing approximation algorithms. We demonstrate the generative leg of the GAL loop through two rounds of algorithmic improvement for Context Extension and KV compression. In both cases, the newly discovered algorithms in each iteration of GAL, address the failure modes identified by the newly generated benchmarks while also maintaining, and  often improving, performance on previously established benchmarks. This behavior illustrates the feedback loop: newly generated benchmarks expose weaknesses in existing methods, which motivates the discovery of improved methods that address these weaknesses without sacrificing performance on existing evaluations.

%% file: sections/related-work.tex
\looseness=-1 \textbf{Auto research and evolution algorithms}
Recent advances in auto research have increasingly used evolutionary search to create autonomous loops that iteratively propose, evaluate, and refine solutions against fixed benchmarks or metrics \citep{karpathy2025autoresearch}. In this paradigm, an LLM is not treated as a one shot solution generator, but as a search operator that proposes mutations or new candidates, while automated evaluation provides the objective signal needed to guide subsequent iterations. AlphaEvolve \citep{novikov2025alphaevolve} exemplifies this approach by combining LLM generated program mutations with evolutionary selection, allowing promising programs to be iteratively improved and complementary ideas to be explored across a population of candidates. GEPA \citep{agrawal2026gepa}, in contrast, uses LLMs to reflect on execution traces and failures to generate targeted mutations, while maintaining a Pareto frontier of candidates that perform well on different subsets of the evaluation data. Related efforts such as autoresearch further demonstrate the potential of autonomous experimentation, where agents repeatedly modify a system, run experiments, and retain improvements based on measured performance. AI Driven Systems Research \citep{cheng2025barbarians} provides a large scale case study of using such automated research agents to tackle challenging systems problems. Collectively, these efforts establish a paradigm in which automated evaluation and evolutionary proposal mechanisms enable LLMs to autonomously navigate large spaces of possible solutions. However, existing auto research systems have primarily used these capabilities to improve a solution against a fixed task, benchmark, or metric. In contrast, we repurpose the same evolutionary machinery to generate adversarial data that targets both the task and the current best solution, enabling the automated discovery of failure modes that can subsequently be used to improve the underlying system.

\textbf{Synthetic data generation}
The field of synthetic data generation is closely related. One of the early influential approaches, Self-Instruct \citep{wang2023self}, showed that an LLM can generate its own instructions, inputs, and outputs, and use the resulting synthetic dataset for instruction tuning. This led to various efforts to evolve the difficulty of synthetic examples and adapt them to different domains \citep{xu2024wizardlm, luo2024wizardcoder, luo2025wizardmath}. Additionally, LLMs have been used to improve the quality of generated data through critique and evaluation \citep{cui2023ultrafeedback, madaan2023self}. A particularly important development has been the use of synthetic data for reasoning distillation. DeepSeek-R1 \citep{guo2025deepseek} demonstrates that a strong reasoning model can generate large numbers of reasoning trajectories that are subsequently used to train smaller models. Collectively, these approaches treat synthetic data generation as a mechanism for constructing increasingly useful training distributions, with generation, evaluation, filtering, and distillation forming a recurring loop. Our work differs in that we use automated search not primarily to generate useful training examples, but to actively generate adversarial examples that target weaknesses of the current best solution.

\looseness=-1 \textbf{Other related works}
There have been several recent and concurrent works that target dataset evolution. BenchEvolver\citep{wu2026benchevolver} proposes a system for evolving coding benchmarks to generate challenging datasets for frontier models. Data-Prompt Co-evolution\citep{lee2026data} evolves test sets and prompts to develop safer and more policy compliant systems. Hacker-Fixer Loops\citep{zhong2026hardening} evolve agentic benchmarks to mitigate reward hacking and ensure that improvements reflect genuine progress. These works are aligned with our approach, with the differences being their areas of application and the domain specific validation used to assess the generated data.

%% file: sections/method_1.tex
In this section, we describe the general formulation of a Generative Adversarial Loop (GAL). Consider a system $\mathcal{S}$, such as an inference system, that takes a
sequence of tokens (e.g., text, images, or other modalities) as input and
produces a sequence as output. The system is parameterized by an artifact
$A$, such as an efficiency algorithm, along with other parameters that
remain fixed throughout the lifetime of the loop. Let $\mathcal{D}$ denote
the set of all valid inputs (or data points), and let $\mathcal{A}$ denote
the set of all valid artifacts that can parameterize $\mathcal{S}$. Note
that the set $\mathcal{D}$ is enumerable, and in systems of interest, such
as LLM inference systems, it is also finite, since the system can only
ingest finite-length strings. If $N$ is the maximum length, in tokens, of
a string the system can ingest, then $D_N$, the set of valid inputs of length $<=N$, is actually a finite set. Let \(\mathcal{E}\) be an evaluation function that determines whether the response of \(\mathcal{S}(A)\) on a given data point is successful. Formally, \(\mathcal{E}: \mathcal{A} \times \mathcal{D} \rightarrow \{0,1\}\),
where \(\mathcal{E}(A,d)=1\) indicates that the system \(\mathcal{S}(A)\) successfully solves data point \(d\in\mathcal{D}\), while \(\mathcal{E}(A,d)=0\) indicates failure. For an artifact \(A\in\mathcal{A}\), we define its coverage over the task space as
\begin{equation}
\mathcal{C}(A) = \{d\in\mathcal{D}\mid \mathcal{E}(A,d)=1\}.
\end{equation}
Thus, \(\mathcal{C}(A)\) represents the subset of the task space on which the artifact \(A\) successfully performs the task. In practice, evaluation is performed on a finite benchmark \(B\subseteq\mathcal{D}\), consisting of a representative set of data points used to measure system performance. For example, \(B\) may comprise long-context benchmarks~\cite{hsieh2024ruler, bai2024longbench, li2023loogle, Lee2024Loft} for KV-cache compression and sparse attention, or prompts from vBench~\citep{huang2024vbench, zheng2025vbench} for sparse video generation. We define the benchmark-level evaluation of an artifact as \(\mathcal{E}(A,B) = |\{d\in B\mid \mathcal{E}(A,d)=1\}|/|B|\).
This quantity measures the fraction of benchmark instances solved by the system parameterized by \(A\). We want to find an algorithm $A^*\in\mathcal{A}$ such that
\begin{equation}
A^* \in \arg\max_{A\in\mathcal{A}} |\mathcal{C}(A)|
\label{eq:objective}
\end{equation}
\subsection{Discriminator and  Generator Agents}
Assume we start with a benchmark $B_0$ and an artifact $A_0$ s.t. $\mathcal{S}(A)$ solves the benchmark $B_0$, i.e., $\mathcal{E}(A_0, B_0) = 1$. We decompose the process of artifact discovery into two phases, executed alternately. In the general case of  $i^{th}$ iteration, we have $B_{i-1}, A_{i-1}$ where $\mathcal{E}(A_{i-1}, B_{i-1}) = 1$

\looseness=-1 \textbf{\redagent{} phase.} 
In the $i^{th}$ iteration of this phase, we seek a data point $b$ on which $A_{i-1}$ fails:
\begin{equation}
    \textrm{\textbf{find} } b \quad \textrm{ s.t. } \quad  b \in \mathcal{D}\setminus\mathcal{C}(A_{i-1})
\end{equation}
Note that, in general, \(\mathcal{C}(A_{i-1})\) can be larger than the benchmark \(B_{i-1}\) used to evaluate \(A_{i-1}\). The discriminator phase seeks to explore the task space beyond the current benchmark by discovering an example that is difficult for the current system \(\mathcal{S}(A_{i-1})\) to solve. Let \(b\in\mathcal{D}\setminus\mathcal{C}(A_{i-1})\) denote such an example. We append this newly discovered example to the existing benchmark \(B_{i-1}\) to obtain an expanded benchmark \(B_i = B_{i-1} \cup \{b\}\), \(B_0 \subset B_1 \subset B_2 \subset \ldots \subseteq \mathcal{D}\).
As the loop proceeds, the benchmark accumulates increasingly challenging examples. In practice, we generalize this process to discover a batch of data points on which \(\mathcal{S}(A_{i-1})\) fails within each discriminator phase.





\looseness=-1 \textbf{\blueagent{} phase.}
This phase seeks an artifact \(A_i\) that solves the expanded benchmark \(B_i\):
\begin{equation}
\textrm{\textbf{find }} A_i
\quad \textrm{s.t.} \quad
\mathcal{E}(A_i,B_i)=1.
\end{equation}
This phase therefore seeks a new artifact that addresses the newly discovered failure cases while preserving correctness on previously discovered examples. In this sense, the generator phase corresponds to the search for improved algorithms or systems, which has traditionally been the focus of evolutionary algorithms and automated research systems. The formulation above presents a simplified, deterministic view in which the correctness of \(\mathcal{S}(A)\) on each data point is binary. The framework can naturally be extended to stochastic systems, such as LLM inference engines, by treating the evaluation outcome as a random variable. Specifically, for a given artifact \(A\) and data point \(d\), we can define \(\mathcal{E}(A,d) \sim \mathrm{Bernoulli}(p_{A,d})\), \(\mathbb{E}[\mathcal{E}(A,B)] = \sum_{d\in B} p_{A,d}/|B|\),
where \(p_{A,d}\) denotes the probability that \(\mathcal{S}(A)\) successfully solves \(d\). The benchmark-level evaluation can then be defined in terms of the expected success rate $\mathbb{E}[\mathcal{E}(A,B)]$.

Since the task space \(\mathcal{D}_N\) is finite, it is guaranteed that the sequence of expanding benchmarks $\{B_i\}$ will eventually reach $\mathcal{D}_N$. Thus, if at every step, our generator is able to find an artifact, we are guaranteed to converge to the true artifact $A^*$ if one exists. This process is illustrated in Figure~\ref{fig:coverage-panels}.

\subsection{Evolution in practice }
\looseness=-1 Since we apply Generative Adversarial Loops to the specific domain of efficiency problems in LLM inference, we have a domain-specific mechanism for both generating candidate examples and verifying their correctness. For long-context problems such as Sparse Attention, KV Compression, and Context Extension, we find it useful to iterate on a program that generates examples rather than directly optimizing the examples themselves. First, this leads to interpretable failure patterns once we identify a candidate point $b$, which would otherwise be difficult to decipher from long-context example. Second, it provides a guarantee of validity by construction. Since the required information is first generated and subsequently embedded into a long context, we can guarantee that the answer is present in the context.

We use a second mechanism to further ensure dataset validity: we evaluate each candidate using unapproximated model inference and require it to achieve good performance on the proposed example. Thus, we modify the definition of the discriminator as
$$
    \textrm{\textbf{find}} \;\;\; b \in \mathcal{D} \;\;\; \textrm{s.t.} \;\;
    \mathbb{E}(\mathcal{E}(\phi, b)) > \tau_0
    \wedge
    \left(
    \mathbb{E}(\mathcal{E}(\phi, b))
    -
    \mathbb{E}(\mathcal{E}(A_{i-1}, b))
    \right) > \tau
$$
where $\mathcal{E}(\phi, b)$, i.e., $A = \phi$, refers to unapproximated LLM inference. Here, $\tau_0$ specifies the minimum performance required from unapproximated inference. This ensures that we do not penalize approximation algorithms on tasks that the unapproximated baseline itself cannot solve, and we then search for data points on which the approximated inference performs worse than the unapproximated inference by at least a gap in quality of $\tau$.

Both the discriminator and generator are search problems, which we propose to solve using approaches in auto-research. To this end, we can use a range of algorithms, including but not limited to GEPA, OpenEvolve, Cursor, and Claude Agents. For large projects, where changes to a single file may depend on many other files across the codebase, Cursor and Claude coding harnesses are effective at providing LLMs with the relevant context. This makes them effective during the generative phase, where successful modifications often require understanding and coordinating changes across multiple parts of a project. We denote a proposer with $P$ and a proposed candidate $c_{t+1} = P(c_t, \mathcal{E}_t, \mathcal{F}_t)$, where $c_t$ is the candidate being searched over at time step $t$ and $\mathcal{F}_t$ is the feedback received via some evaluator. We iterate in a loop multiple times until the proposer finds a valid next candidate.

\input{tex_figures/coverage_panel}

\textbf{\redagent{} and \blueagent{}}
At each iteration $i$, given the current benchmark set $B_{i-1}$ on which algorithm $A_{i-1}$ succeeds, \redagent{} uses proposer $P$ to generate a new data point $b = P(B_{i-1}, \mathcal{E}_{i-1}, \mathcal{F}_{i-1}),$
and evaluate $A_{i-1}$ on $b$. We keep generating a data point till we find one that $A_{i-1}$ cannot solve. This $b$ is then appended to $B_{i-1}$ to get $B_i$
Then, \blueagent{} takes over and uses proposer $P$ to generate a new algorithm $A_i$ which not only solves $b$ but also solves $B_{i-1}$.

%% file: tex_figures/coverage_panel.tex
\begin{figure*}[t]
\centering
\begin{minipage}[c]{0.38\textwidth}
\centering
\resizebox{\textwidth}{!}{%
\begin{tikzpicture}[x=2.2cm, y=1.870cm, font=\small]

  \def\taskspace{(1.420,0.000) (1.414,0.046) (1.407,0.092) (1.397,0.138) (1.386,0.182) (1.373,0.227) (1.359,0.270) (1.343,0.316) (1.327,0.359) (1.311,0.401) (1.294,0.442) (1.276,0.483) (1.258,0.524) (1.239,0.568) (1.220,0.608) (1.202,0.649) (1.183,0.690) (1.164,0.731) (1.144,0.772) (1.123,0.816) (1.102,0.857) (1.081,0.899) (1.059,0.941) (1.035,0.982) (1.010,1.023) (0.984,1.064) (0.954,1.107) (0.924,1.147) (0.893,1.185) (0.859,1.222) (0.824,1.257) (0.787,1.290) (0.745,1.323) (0.703,1.351) (0.661,1.377) (0.616,1.400) (0.570,1.420) (0.523,1.437) (0.472,1.451) (0.423,1.462) (0.373,1.469) (0.323,1.473) (0.273,1.474) (0.223,1.472) (0.174,1.468) (0.121,1.460) (0.073,1.451) (0.025,1.440) (-0.022,1.427) (-0.068,1.412) (-0.113,1.396) (-0.160,1.378) (-0.203,1.360) (-0.246,1.341) (-0.287,1.322) (-0.328,1.303) (-0.368,1.283) (-0.410,1.262) (-0.449,1.242) (-0.487,1.222) (-0.525,1.201) (-0.563,1.181) (-0.601,1.160) (-0.638,1.139) (-0.677,1.115) (-0.713,1.093) (-0.749,1.070) (-0.785,1.046) (-0.820,1.021) (-0.854,0.995) (-0.890,0.967) (-0.923,0.939) (-0.955,0.910) (-0.986,0.880) (-1.016,0.849) (-1.045,0.817) (-1.075,0.781) (-1.102,0.747) (-1.128,0.712) (-1.152,0.675) (-1.176,0.639) (-1.199,0.601) (-1.220,0.563) (-1.243,0.521) (-1.263,0.482) (-1.282,0.441) (-1.300,0.401) (-1.318,0.359) (-1.335,0.318) (-1.353,0.272) (-1.369,0.229) (-1.385,0.185) (-1.400,0.141) (-1.414,0.096) (-1.427,0.050) (-1.440,0.000) (-1.451,-0.047) (-1.461,-0.096) (-1.469,-0.145) (-1.475,-0.194) (-1.480,-0.244) (-1.482,-0.295) (-1.482,-0.349) (-1.479,-0.400) (-1.473,-0.450) (-1.465,-0.501) (-1.453,-0.551) (-1.439,-0.600) (-1.420,-0.651) (-1.399,-0.698) (-1.375,-0.743) (-1.348,-0.786) (-1.318,-0.827) (-1.285,-0.867) (-1.248,-0.906) (-1.210,-0.941) (-1.170,-0.973) (-1.129,-1.003) (-1.086,-1.031) (-1.042,-1.056) (-0.997,-1.079) (-0.948,-1.101) (-0.902,-1.119) (-0.856,-1.136) (-0.810,-1.152) (-0.764,-1.165) (-0.719,-1.178) (-0.670,-1.191) (-0.626,-1.202) (-0.582,-1.213) (-0.538,-1.223) (-0.495,-1.233) (-0.452,-1.243) (-0.407,-1.254) (-0.365,-1.264) (-0.323,-1.274) (-0.282,-1.284) (-0.240,-1.294) (-0.198,-1.304) (-0.155,-1.313) (-0.110,-1.323) (-0.067,-1.331) (-0.023,-1.338) (0.021,-1.345) (0.065,-1.350) (0.109,-1.353) (0.157,-1.356) (0.203,-1.356) (0.248,-1.354) (0.293,-1.351) (0.338,-1.345) (0.383,-1.337) (0.431,-1.326) (0.475,-1.314) (0.518,-1.299) (0.561,-1.283) (0.603,-1.264) (0.644,-1.244) (0.685,-1.222) (0.726,-1.197) (0.765,-1.172) (0.802,-1.146) (0.839,-1.118) (0.874,-1.089) (0.909,-1.059) (0.945,-1.027) (0.978,-0.995) (1.010,-0.963) (1.042,-0.930) (1.073,-0.896) (1.103,-0.862) (1.134,-0.824) (1.163,-0.788) (1.191,-0.751) (1.217,-0.714) (1.243,-0.675) (1.268,-0.636) (1.291,-0.595) (1.314,-0.551) (1.335,-0.509) (1.353,-0.466) (1.370,-0.422) (1.384,-0.377) (1.397,-0.332) (1.408,-0.283) (1.416,-0.237) (1.421,-0.190) (1.425,-0.143) (1.425,-0.097) (1.424,-0.050)}
  \def\benchset{(0.956,0.000) (0.979,0.032) (1.001,0.066) (1.020,0.100) (1.037,0.137) (1.051,0.173) (1.060,0.211) (1.064,0.251) (1.063,0.287) (1.056,0.323) (1.043,0.356) (1.024,0.388) (0.999,0.416) (0.967,0.443) (0.933,0.465) (0.894,0.483) (0.853,0.497) (0.809,0.508) (0.764,0.516) (0.716,0.520) (0.671,0.522) (0.627,0.521) (0.584,0.519) (0.543,0.515) (0.504,0.511) (0.467,0.506) (0.431,0.500) (0.399,0.495) (0.369,0.490) (0.342,0.486) (0.316,0.482) (0.292,0.479) (0.269,0.477) (0.248,0.476) (0.228,0.476) (0.210,0.478) (0.193,0.480) (0.176,0.483) (0.159,0.489) (0.143,0.495) (0.127,0.502) (0.112,0.510) (0.096,0.520) (0.080,0.531) (0.064,0.542) (0.046,0.555) (0.029,0.568) (0.010,0.582) (-0.009,0.595) (-0.029,0.608) (-0.050,0.621) (-0.074,0.634) (-0.096,0.645) (-0.120,0.655) (-0.144,0.663) (-0.168,0.669) (-0.193,0.673) (-0.219,0.675) (-0.244,0.674) (-0.267,0.670) (-0.291,0.664) (-0.313,0.656) (-0.334,0.645) (-0.354,0.632) (-0.373,0.615) (-0.390,0.597) (-0.405,0.578) (-0.418,0.557) (-0.429,0.535) (-0.439,0.512) (-0.448,0.486) (-0.454,0.462) (-0.459,0.438) (-0.463,0.414) (-0.467,0.390) (-0.469,0.367) (-0.472,0.343) (-0.474,0.321) (-0.476,0.300) (-0.478,0.280) (-0.481,0.261) (-0.484,0.243) (-0.489,0.225) (-0.494,0.207) (-0.500,0.191) (-0.507,0.175) (-0.515,0.159) (-0.523,0.143) (-0.532,0.127) (-0.542,0.109) (-0.552,0.092) (-0.562,0.075) (-0.571,0.058) (-0.580,0.039) (-0.589,0.021) (-0.597,0.000) (-0.603,-0.020) (-0.608,-0.040) (-0.611,-0.060) (-0.613,-0.081) (-0.614,-0.101) (-0.612,-0.122) (-0.609,-0.143) (-0.604,-0.163) (-0.598,-0.183) (-0.590,-0.202) (-0.581,-0.220) (-0.571,-0.238) (-0.559,-0.256) (-0.547,-0.273) (-0.534,-0.288) (-0.521,-0.304) (-0.507,-0.318) (-0.493,-0.333) (-0.479,-0.348) (-0.465,-0.362) (-0.451,-0.375) (-0.438,-0.389) (-0.424,-0.403) (-0.411,-0.416) (-0.398,-0.430) (-0.384,-0.445) (-0.371,-0.460) (-0.357,-0.474) (-0.343,-0.488) (-0.329,-0.502) (-0.315,-0.516) (-0.299,-0.531) (-0.283,-0.544) (-0.267,-0.556) (-0.250,-0.567) (-0.232,-0.577) (-0.213,-0.586) (-0.193,-0.594) (-0.173,-0.599) (-0.153,-0.603) (-0.133,-0.606) (-0.112,-0.606) (-0.092,-0.604) (-0.071,-0.601) (-0.050,-0.596) (-0.030,-0.589) (-0.010,-0.581) (0.009,-0.572) (0.027,-0.562) (0.045,-0.552) (0.063,-0.540) (0.079,-0.529) (0.095,-0.518) (0.110,-0.507) (0.125,-0.497) (0.140,-0.487) (0.155,-0.478) (0.170,-0.471) (0.185,-0.464) (0.201,-0.459) (0.217,-0.455) (0.234,-0.452) (0.252,-0.450) (0.272,-0.449) (0.292,-0.448) (0.313,-0.448) (0.336,-0.447) (0.359,-0.447) (0.383,-0.446) (0.409,-0.445) (0.435,-0.442) (0.460,-0.439) (0.486,-0.434) (0.512,-0.428) (0.538,-0.420) (0.564,-0.410) (0.588,-0.399) (0.612,-0.386) (0.634,-0.371) (0.655,-0.356) (0.675,-0.339) (0.695,-0.320) (0.714,-0.300) (0.732,-0.279) (0.750,-0.258) (0.767,-0.236) (0.785,-0.214) (0.803,-0.191) (0.823,-0.166) (0.843,-0.141) (0.864,-0.116) (0.885,-0.089) (0.908,-0.062) (0.931,-0.033)}
  \def\bone{(0.631,0.000) (0.635,0.021) (0.638,0.042) (0.639,0.063) (0.638,0.084) (0.635,0.105) (0.630,0.125) (0.623,0.147) (0.614,0.166) (0.606,0.185) (0.596,0.204) (0.587,0.222) (0.577,0.240) (0.566,0.260) (0.556,0.277) (0.545,0.295) (0.534,0.312) (0.522,0.328) (0.510,0.344) (0.496,0.360) (0.482,0.375) (0.467,0.388) (0.451,0.401) (0.434,0.412) (0.416,0.422) (0.398,0.431) (0.378,0.438) (0.357,0.443) (0.337,0.447) (0.316,0.449) (0.295,0.450) (0.275,0.451) (0.253,0.450) (0.234,0.449) (0.215,0.447) (0.196,0.446) (0.178,0.444) (0.161,0.443) (0.143,0.441) (0.127,0.440) (0.111,0.439) (0.096,0.438) (0.081,0.437) (0.066,0.437) (0.052,0.438) (0.036,0.439) (0.022,0.441) (0.008,0.444) (-0.007,0.447) (-0.022,0.451) (-0.037,0.456) (-0.054,0.461) (-0.070,0.466) (-0.086,0.472) (-0.104,0.477) (-0.121,0.482) (-0.140,0.487) (-0.160,0.492) (-0.179,0.495) (-0.199,0.498) (-0.219,0.500) (-0.239,0.501) (-0.259,0.501) (-0.279,0.499) (-0.301,0.496) (-0.320,0.491) (-0.339,0.485) (-0.358,0.477) (-0.375,0.467) (-0.391,0.456) (-0.406,0.442) (-0.419,0.426) (-0.429,0.409) (-0.437,0.390) (-0.444,0.371) (-0.449,0.351) (-0.453,0.329) (-0.456,0.309) (-0.459,0.290) (-0.462,0.271) (-0.465,0.253) (-0.469,0.235) (-0.472,0.218) (-0.476,0.200) (-0.481,0.183) (-0.485,0.167) (-0.489,0.151) (-0.492,0.134) (-0.494,0.117) (-0.495,0.100) (-0.496,0.083) (-0.495,0.066) (-0.494,0.050) (-0.492,0.033) (-0.490,0.017) (-0.489,0.000) (-0.487,-0.016) (-0.485,-0.032) (-0.483,-0.048) (-0.482,-0.063) (-0.480,-0.079) (-0.478,-0.095) (-0.476,-0.112) (-0.474,-0.128) (-0.471,-0.144) (-0.468,-0.160) (-0.465,-0.176) (-0.461,-0.192) (-0.456,-0.209) (-0.450,-0.224) (-0.444,-0.240) (-0.437,-0.255) (-0.429,-0.269) (-0.420,-0.284) (-0.410,-0.298) (-0.400,-0.311) (-0.389,-0.324) (-0.377,-0.335) (-0.365,-0.346) (-0.352,-0.357) (-0.339,-0.366) (-0.324,-0.376) (-0.309,-0.384) (-0.295,-0.391) (-0.280,-0.397) (-0.264,-0.403) (-0.249,-0.408) (-0.232,-0.413) (-0.217,-0.416) (-0.201,-0.420) (-0.186,-0.423) (-0.171,-0.426) (-0.156,-0.428) (-0.140,-0.431) (-0.125,-0.433) (-0.111,-0.436) (-0.096,-0.439) (-0.082,-0.441) (-0.067,-0.445) (-0.053,-0.448) (-0.038,-0.452) (-0.023,-0.455) (-0.008,-0.459) (0.007,-0.463) (0.022,-0.467) (0.038,-0.471) (0.055,-0.474) (0.071,-0.476) (0.087,-0.477) (0.103,-0.475) (0.119,-0.472) (0.134,-0.467) (0.150,-0.462) (0.165,-0.456) (0.180,-0.451) (0.195,-0.447) (0.211,-0.443) (0.228,-0.440) (0.245,-0.438) (0.265,-0.436) (0.284,-0.435) (0.304,-0.434) (0.324,-0.432) (0.345,-0.430) (0.367,-0.427) (0.389,-0.423) (0.409,-0.416) (0.428,-0.408) (0.446,-0.398) (0.461,-0.385) (0.476,-0.372) (0.489,-0.356) (0.501,-0.340) (0.512,-0.323) (0.522,-0.306) (0.532,-0.289) (0.541,-0.271) (0.549,-0.253) (0.557,-0.234) (0.565,-0.215) (0.572,-0.197) (0.579,-0.178) (0.586,-0.160) (0.592,-0.141) (0.599,-0.120) (0.605,-0.101) (0.611,-0.082) (0.616,-0.062) (0.622,-0.042) (0.627,-0.022)}
  \def\cova{(0.744,0.000) (0.730,0.024) (0.716,0.047) (0.703,0.069) (0.689,0.091) (0.676,0.112) (0.664,0.132) (0.651,0.153) (0.640,0.173) (0.629,0.192) (0.619,0.211) (0.608,0.230) (0.598,0.249) (0.588,0.269) (0.577,0.288) (0.567,0.306) (0.556,0.324) (0.544,0.342) (0.532,0.359) (0.517,0.376) (0.503,0.391) (0.488,0.406) (0.472,0.420) (0.456,0.432) (0.438,0.444) (0.420,0.454) (0.400,0.464) (0.380,0.472) (0.361,0.479) (0.341,0.485) (0.322,0.490) (0.302,0.495) (0.281,0.500) (0.262,0.504) (0.244,0.508) (0.226,0.513) (0.208,0.518) (0.190,0.523) (0.172,0.530) (0.155,0.537) (0.139,0.546) (0.122,0.556) (0.105,0.567) (0.088,0.579) (0.070,0.592) (0.050,0.607) (0.031,0.623) (0.011,0.638) (-0.010,0.655) (-0.032,0.672) (-0.056,0.688) (-0.082,0.705) (-0.108,0.721) (-0.135,0.735) (-0.162,0.748) (-0.191,0.759) (-0.220,0.768) (-0.252,0.776) (-0.282,0.780) (-0.312,0.781) (-0.341,0.780) (-0.370,0.775) (-0.398,0.768) (-0.424,0.758) (-0.451,0.743) (-0.474,0.727) (-0.496,0.708) (-0.516,0.687) (-0.533,0.664) (-0.549,0.640) (-0.563,0.612) (-0.574,0.584) (-0.584,0.557) (-0.592,0.528) (-0.599,0.500) (-0.604,0.472) (-0.609,0.443) (-0.614,0.416) (-0.618,0.390) (-0.622,0.364) (-0.626,0.340) (-0.630,0.316) (-0.636,0.293) (-0.642,0.269) (-0.648,0.247) (-0.656,0.226) (-0.664,0.204) (-0.672,0.183) (-0.681,0.162) (-0.690,0.139) (-0.698,0.117) (-0.706,0.095) (-0.713,0.072) (-0.719,0.049) (-0.724,0.025) (-0.728,0.000) (-0.729,-0.024) (-0.729,-0.048) (-0.726,-0.072) (-0.722,-0.095) (-0.716,-0.118) (-0.707,-0.141) (-0.697,-0.164) (-0.685,-0.185) (-0.671,-0.205) (-0.657,-0.225) (-0.641,-0.243) (-0.624,-0.260) (-0.606,-0.278) (-0.588,-0.293) (-0.570,-0.308) (-0.553,-0.322) (-0.535,-0.336) (-0.518,-0.349) (-0.500,-0.363) (-0.484,-0.377) (-0.469,-0.390) (-0.454,-0.403) (-0.439,-0.417) (-0.425,-0.431) (-0.412,-0.446) (-0.398,-0.462) (-0.385,-0.478) (-0.372,-0.494) (-0.359,-0.511) (-0.346,-0.528) (-0.333,-0.546) (-0.318,-0.564) (-0.303,-0.581) (-0.287,-0.598) (-0.271,-0.615) (-0.253,-0.630) (-0.235,-0.645) (-0.214,-0.659) (-0.194,-0.671) (-0.173,-0.681) (-0.151,-0.688) (-0.129,-0.694) (-0.106,-0.698) (-0.083,-0.700) (-0.058,-0.700) (-0.035,-0.697) (-0.012,-0.692) (0.010,-0.686) (0.033,-0.679) (0.054,-0.670) (0.077,-0.659) (0.097,-0.649) (0.117,-0.638) (0.136,-0.627) (0.155,-0.617) (0.174,-0.607) (0.194,-0.597) (0.213,-0.589) (0.232,-0.582) (0.252,-0.577) (0.273,-0.572) (0.294,-0.568) (0.317,-0.565) (0.342,-0.563) (0.367,-0.562) (0.393,-0.561) (0.420,-0.559) (0.448,-0.558) (0.476,-0.555) (0.508,-0.552) (0.537,-0.547) (0.567,-0.541) (0.597,-0.533) (0.626,-0.523) (0.653,-0.511) (0.682,-0.495) (0.707,-0.479) (0.730,-0.460) (0.750,-0.440) (0.769,-0.417) (0.785,-0.393) (0.798,-0.368) (0.808,-0.339) (0.815,-0.311) (0.820,-0.282) (0.821,-0.253) (0.820,-0.224) (0.817,-0.194) (0.811,-0.163) (0.803,-0.134) (0.794,-0.106) (0.783,-0.079) (0.771,-0.052) (0.758,-0.026)}
  \def\covb{(1.070,0.000) (1.095,0.036) (1.119,0.073) (1.141,0.112) (1.159,0.153) (1.174,0.194) (1.184,0.236) (1.189,0.280) (1.187,0.321) (1.179,0.360) (1.164,0.398) (1.143,0.433) (1.115,0.465) (1.079,0.495) (1.040,0.518) (0.996,0.538) (0.950,0.554) (0.900,0.565) (0.850,0.573) (0.795,0.578) (0.744,0.579) (0.694,0.577) (0.646,0.574) (0.599,0.569) (0.555,0.563) (0.514,0.556) (0.473,0.548) (0.437,0.542) (0.403,0.535) (0.372,0.529) (0.344,0.524) (0.317,0.520) (0.291,0.516) (0.268,0.514) (0.246,0.513) (0.226,0.514) (0.208,0.518) (0.190,0.523) (0.172,0.530) (0.155,0.537) (0.139,0.546) (0.122,0.556) (0.105,0.567) (0.088,0.579) (0.070,0.592) (0.050,0.607) (0.031,0.623) (0.011,0.638) (-0.010,0.655) (-0.032,0.672) (-0.056,0.688) (-0.082,0.705) (-0.108,0.721) (-0.135,0.735) (-0.162,0.748) (-0.191,0.759) (-0.220,0.768) (-0.252,0.776) (-0.282,0.780) (-0.312,0.781) (-0.341,0.780) (-0.370,0.775) (-0.398,0.768) (-0.424,0.758) (-0.451,0.743) (-0.474,0.727) (-0.496,0.708) (-0.516,0.687) (-0.533,0.664) (-0.549,0.640) (-0.563,0.612) (-0.574,0.584) (-0.584,0.557) (-0.592,0.528) (-0.599,0.500) (-0.604,0.472) (-0.609,0.443) (-0.614,0.416) (-0.618,0.390) (-0.622,0.364) (-0.626,0.340) (-0.630,0.316) (-0.636,0.293) (-0.642,0.269) (-0.648,0.247) (-0.656,0.226) (-0.664,0.204) (-0.672,0.183) (-0.681,0.162) (-0.690,0.139) (-0.698,0.117) (-0.706,0.095) (-0.713,0.072) (-0.719,0.049) (-0.724,0.025) (-0.728,0.000) (-0.729,-0.024) (-0.729,-0.048) (-0.726,-0.072) (-0.722,-0.095) (-0.716,-0.118) (-0.707,-0.141) (-0.697,-0.164) (-0.685,-0.185) (-0.671,-0.205) (-0.657,-0.225) (-0.641,-0.243) (-0.624,-0.260) (-0.606,-0.278) (-0.588,-0.293) (-0.570,-0.308) (-0.553,-0.322) (-0.535,-0.336) (-0.518,-0.349) (-0.500,-0.363) (-0.484,-0.377) (-0.469,-0.390) (-0.454,-0.403) (-0.439,-0.417) (-0.425,-0.431) (-0.412,-0.446) (-0.398,-0.462) (-0.385,-0.478) (-0.372,-0.494) (-0.359,-0.511) (-0.346,-0.528) (-0.333,-0.546) (-0.318,-0.564) (-0.303,-0.581) (-0.287,-0.598) (-0.271,-0.615) (-0.253,-0.630) (-0.235,-0.645) (-0.214,-0.659) (-0.194,-0.671) (-0.173,-0.681) (-0.151,-0.688) (-0.129,-0.694) (-0.106,-0.698) (-0.083,-0.700) (-0.058,-0.700) (-0.035,-0.697) (-0.012,-0.692) (0.010,-0.686) (0.033,-0.679) (0.054,-0.670) (0.077,-0.659) (0.097,-0.649) (0.117,-0.638) (0.136,-0.627) (0.155,-0.617) (0.174,-0.607) (0.194,-0.597) (0.213,-0.589) (0.232,-0.582) (0.252,-0.577) (0.273,-0.572) (0.294,-0.568) (0.317,-0.565) (0.342,-0.563) (0.367,-0.562) (0.393,-0.561) (0.420,-0.559) (0.448,-0.558) (0.476,-0.555) (0.508,-0.552) (0.537,-0.547) (0.567,-0.541) (0.597,-0.533) (0.626,-0.523) (0.653,-0.511) (0.682,-0.495) (0.707,-0.479) (0.730,-0.460) (0.750,-0.440) (0.769,-0.417) (0.785,-0.393) (0.798,-0.368) (0.813,-0.341) (0.831,-0.317) (0.849,-0.292) (0.867,-0.267) (0.886,-0.241) (0.905,-0.215) (0.926,-0.186) (0.947,-0.159) (0.970,-0.130) (0.993,-0.100) (1.018,-0.069) (1.043,-0.036)}
  \def\covn{(1.360,0.000) (1.353,0.044) (1.344,0.088) (1.333,0.131) (1.321,0.174) (1.307,0.216) (1.291,0.257) (1.273,0.300) (1.255,0.339) (1.236,0.378) (1.217,0.416) (1.198,0.454) (1.178,0.491) (1.157,0.530) (1.138,0.567) (1.118,0.604) (1.099,0.641) (1.079,0.678) (1.060,0.715) (1.039,0.755) (1.019,0.792) (0.998,0.830) (0.977,0.869) (0.956,0.907) (0.933,0.945) (0.909,0.984) (0.883,1.024) (0.856,1.062) (0.828,1.099) (0.798,1.134) (0.766,1.169) (0.733,1.202) (0.695,1.235) (0.658,1.264) (0.619,1.291) (0.579,1.315) (0.537,1.337) (0.494,1.356) (0.446,1.373) (0.401,1.386) (0.354,1.396) (0.308,1.402) (0.261,1.406) (0.214,1.408) (0.166,1.406) (0.116,1.401) (0.070,1.395) (0.024,1.386) (-0.021,1.375) (-0.065,1.362) (-0.109,1.348) (-0.155,1.331) (-0.196,1.314) (-0.237,1.296) (-0.277,1.277) (-0.316,1.258) (-0.355,1.237) (-0.395,1.215) (-0.432,1.194) (-0.468,1.172) (-0.503,1.150) (-0.537,1.127) (-0.571,1.104) (-0.605,1.080) (-0.639,1.054) (-0.671,1.029) (-0.702,1.003) (-0.732,0.976) (-0.762,0.949) (-0.790,0.921) (-0.820,0.891) (-0.846,0.861) (-0.872,0.831) (-0.896,0.800) (-0.920,0.768) (-0.942,0.736) (-0.965,0.701) (-0.985,0.668) (-1.005,0.634) (-1.023,0.600) (-1.041,0.565) (-1.059,0.531) (-1.075,0.496) (-1.093,0.458) (-1.109,0.423) (-1.126,0.388) (-1.142,0.352) (-1.158,0.316) (-1.174,0.279) (-1.192,0.240) (-1.209,0.202) (-1.226,0.164) (-1.243,0.125) (-1.260,0.085) (-1.276,0.045) (-1.294,0.000) (-1.310,-0.043) (-1.325,-0.087) (-1.339,-0.132) (-1.352,-0.178) (-1.363,-0.225) (-1.373,-0.273) (-1.380,-0.325) (-1.385,-0.374) (-1.387,-0.424) (-1.386,-0.474) (-1.382,-0.524) (-1.375,-0.573) (-1.363,-0.625) (-1.349,-0.673) (-1.331,-0.719) (-1.310,-0.764) (-1.286,-0.807) (-1.258,-0.849) (-1.225,-0.890) (-1.192,-0.927) (-1.156,-0.961) (-1.117,-0.993) (-1.076,-1.022) (-1.034,-1.048) (-0.991,-1.072) (-0.943,-1.094) (-0.897,-1.113) (-0.851,-1.129) (-0.804,-1.143) (-0.758,-1.156) (-0.711,-1.166) (-0.662,-1.176) (-0.616,-1.184) (-0.571,-1.191) (-0.527,-1.197) (-0.483,-1.202) (-0.439,-1.207) (-0.394,-1.211) (-0.351,-1.215) (-0.310,-1.219) (-0.268,-1.222) (-0.227,-1.226) (-0.186,-1.228) (-0.146,-1.231) (-0.102,-1.233) (-0.062,-1.235) (-0.022,-1.236) (0.019,-1.236) (0.059,-1.235) (0.100,-1.233) (0.143,-1.230) (0.183,-1.226) (0.223,-1.220) (0.263,-1.214) (0.303,-1.205) (0.343,-1.195) (0.385,-1.183) (0.423,-1.171) (0.461,-1.156) (0.499,-1.141) (0.536,-1.124) (0.573,-1.106) (0.609,-1.087) (0.647,-1.065) (0.681,-1.044) (0.716,-1.022) (0.750,-0.999) (0.783,-0.976) (0.816,-0.952) (0.851,-0.925) (0.884,-0.899) (0.916,-0.873) (0.948,-0.846) (0.980,-0.818) (1.011,-0.790) (1.044,-0.758) (1.074,-0.728) (1.104,-0.696) (1.133,-0.664) (1.161,-0.630) (1.188,-0.595) (1.214,-0.560) (1.240,-0.520) (1.262,-0.481) (1.283,-0.442) (1.302,-0.401) (1.318,-0.359) (1.333,-0.317) (1.345,-0.271) (1.354,-0.227) (1.361,-0.182) (1.364,-0.137) (1.365,-0.092) (1.364,-0.048)}

  \fill[blueteam!7]  plot[smooth cycle] coordinates {\covn};
  \fill[blueteam!19] plot[smooth cycle] coordinates {\covb};
  \fill[blueteam!26] plot[smooth cycle] coordinates {\cova};

  \fill[redteam!45] plot[smooth cycle] coordinates {\benchset};
  \fill[redteam!16] plot[smooth cycle] coordinates {\bone};

  \draw[gray!45, dashed, line width=0.5pt]
    plot[smooth cycle] coordinates {\taskspace};
  \draw[blueteam!40, dashed, line width=0.7pt]
    plot[smooth cycle] coordinates {\covn};
  \draw[blueteam!65, line width=0.6pt] plot[smooth cycle] coordinates {\covb};
  \draw[redteam, line width=0.85pt]    plot[smooth cycle] coordinates {\benchset};
  \draw[redteam!65, line width=0.6pt]  plot[smooth cycle] coordinates {\bone};
  \draw[blueteam, line width=0.85pt]   plot[smooth cycle] coordinates {\cova};

  \node[gray!70, font=\footnotesize] at (-0.156,1.480) {$\mathcal{D}$};
  \node[blueteam, font=\footnotesize] at (-0.211,-0.580) {$A_1$};
  \node[blueteam, font=\footnotesize] at (0.918,-0.371) {$A_2$};
  \node[blueteam, font=\footnotesize] at (0.542,-0.939) {$A_n$};
  \node[redteam, font=\footnotesize] at (0.851,0.293) {$b_2$};
  \node[redteam, font=\footnotesize] at (-0.205,-0.095) {$b_1$};

\end{tikzpicture}}
\end{minipage}\hfill
\begin{minipage}[c]{0.59\textwidth}
\begin{algorithm}[H]
\caption{Local Optimization}
\label{alg:loop}
\begin{algorithmic}[1]
  \Require initial algorithm $A_0$, initial benchmark $B_0 \subset \mathcal{D}$
  
  \For{$i = 1, 2, \dots, n$}
  
    $ b \gets \textrm{\textbf{find} } b \quad \textrm{ s.t. } \quad  b \in \mathcal{D}\setminus\mathcal{C}(A_{i-1})$
    \State ${{B_i}} \gets
           {{B_{i-1}}} \cup \{b\}$
    \State $A_i \gets$ find $\mathcal{A} \textrm{ s.t. } \quad  \mathcal{E}(A_i, B_{i}) = 1$
  \EndFor
  \State \Return $A_n$
\end{algorithmic}
\end{algorithm}
\end{minipage}
\caption{Visualization and the iterative algorithm. The red area represents the benchmark set and blue area is the artifact coverage. After the first iteration $\mathcal{C}(A_1)$ contains $b_1$, so that benchmark is solved. The Discriminator then extends the benchmark with $b_2$ to form $B_1$, which is no longer contained in $\mathcal{C}(A_1)$. The Generator then answers with $A_2$ which covers the new set. After enough number of iterations, the cover of algorithm will approach the dataset $\mathcal{D_N}$.}
\label{fig:coverage-panels}
\end{figure*}

%% file: sections/case_studies_compact.tex
\input{tex_figures/box1}
\looseness=-1 In this section, we describe the problem domains we target, the algorithms we use to represent the state of the art in each domain, and representative adversarial examples discovered by the GAL discriminator, along with their interpretation.

\textbf{\textsc{Sparse Video Generation}}

State-of-the-art video generation typically relies on diffusion models, which iteratively refine a video over multiple steps. Each step requires computation over the full sequence of video tokens, making generation increasingly expensive as resolution and duration grow. As a result, even a few seconds of high-resolution video can take tens of minutes to generate on high-end GPUs. \emph{Sparse prefill} addresses this by reducing attention computation while preserving generation quality. One state-of-the-art approach is SVG (Sparse Video Generation)~\citep{xi2025sparse}. SVG identifies two distinct attention patterns in video - \emph{spatial} and \emph{temporal} and applies different sparsity patterns to each head type. It substantially reduces computation while maintaining quality close to dense generation, reporting PSNR values of approximately 28--29, SSIM around 0.9, and LPIPS around 0.1--0.12, with comparable or sometimes better image quality and temporal consistency.

\textbf{Example Adversary:}
GAL's discriminator identified adversarial prompts that cause SVG to deviate substantially from dense video generation. One such example is shown in Figure~\ref{fig:svg-box}. For this prompt, the relative PSNR and SSIM are substantially lower, while LPIPS is higher, indicating a larger deviation from the video produced by the dense model. The adversarial prompt requests a high-resolution video involving complex interactions of light, creating a challenging generation setting. Such prompts appear to expose a weakness in the sparsity patterns used by SVG: attention that can be safely omitted for typical video-generation prompts may become important when the scene contains fine-grained spatial structure and complex temporal interactions. Thus, while SVG performs well on standard benchmarks, the adversarial example demonstrates that its approximation can breakdown under specific, systematically generated inputs.

\input{tex_figures/box2}

\textbf{\textsc{Context Extension}}

Extending a model's context window is computationally expensive, motivating training-free methods that extend the effective context beyond the native limit. Dual Chunk Attention (DCA)~\citep{an2024training} is a state-of-the-art approach, natively supported in vLLM~\citep{kwon2023efficient}, with particularly strong results on Qwen models. DCA addresses out-of-distribution relative positions by remapping token positions based on their distance from the query. As illustrated in Figure~\ref{fig:dca-box}, tokens from distant chunks are mapped to a shared local range $0$ to $C$, while nearby chunks are positioned immediately preceding the query, where $C$ is the chunk size. DCA typically enables $2\times$--$4\times$ context extension while preserving perplexity and retrieval performance, with some results extending beyond this range.

\textbf{Example Adversary:}
GAL's discriminator identified adversarial patterns that cause DCA to fail substantially. One such pattern is shown in Figure~\ref{fig:dca-box}. The discriminator generates a program parameterized by the context length. At the native context length, the base model successfully solves the problem. However, when the same problem is extended to $4\times$ the native context window, the DCA-enabled model fails. The adversarial pattern exploits the interaction between DCA's positional remapping and the location of the retrieval information. Specifically, the relevant retrieval information is placed at the boundary of a chunk. Although DCA preserves relative positioning locally within each chunk, its remapping can cause tokens belonging to the same retrieval pattern to be scattered across different relative positions when the information lies near a chunk boundary. For a distant query, this scattering changes the positional structure of the retrieval information relative to the query. As a result, the model may fail to recognize and retrieve the relevant information, despite the information itself being present in the context.

\textbf{\textsc{KV Compression}}

Long-context autoregressive models face a memory bottleneck in the form of the KV cache, which stores per-token embeddings to reduce generation-time computation. Since its size grows linearly with context length, large KV caches can limit batch sizes or require CPU offloading, reducing serving throughput. KV Compression addresses this by reducing the cache's memory footprint. KVZip~\mbox{\citep{kim2026kvzip}} and its lightweight variant FastKVZip~\citep{kim2026fast} are state-of-the-art approaches that retain a small subset of important keys and values to approximate the full context. They achieve up to $4\times$ compression while maintaining performance on retrieval, contextual QA, and related tasks.

\textbf{Example Adversary:} GAL's discriminator identified adversarial patterns that cause FastKVZip to fail substantially. One such pattern is shown in Figure~\ref{fig:kvpress-box}. As we can see, FastKVZip fails significantly, dropping the accuracy score from 0.93 for the uncompressed model to 0.35 for the compressed model. The adversarial pattern exploits compression errors by constructing a fresh ledger and a stale ledger of key-value pairs. The fresh ledger is placed mid-context, while the stale ledger appears at the end with each entry explicitly marked stale. The query specifically asks for the fresh ledger. When the context comprises news or articles that are time sensitive, this can be a practical occurrence where stale information is also present in the context.

\textbf{\textsc{Sparse Decoding}}

\looseness=-1 Autoregressive LLM decoding is memory-bound, primarily due to  reading the large KV cache from GPU memory, limiting token throughput. Sparse attention reduces this memory movement by attending to a subset of tokens, making token selection critical. PQCache~\citep{zhang2025pqcache} is a state-of-the-art sparse attention method that selects relevant tokens using a compact product-quantized representation of the KV cache. It achieves top performance on sparse attention leaderboard for Qwen models~\citep{skylight}. We therefore use Qwen3-4B-Instruct-2507 with PQCache at $2\%$ sparsity.
\input{utility/cost_plots_combined}

\textbf{Example Adversary:} GAL's discriminator identifies adversarial patterns that cause PQCache to fail substantially. One such pattern is shown in Figure~\ref{fig:pqcache-box}. As we can see, PQCache performs significantly worse on this pattern. The adversarial pattern exploits the fact that the nature of most attention is such that it naturally pays more attention to sink tokens (i.e., at the start of the context) and local tokens (near the query). When important keys are embedded in the middle, the lost in the middle phenomenon is observed. Newer models, such as Qwen, after training for long contexts, have largely solved the lost in the middle problem, but when sparsified, the phenomenon shows up again. At higher sparsities (e.g., 5\% or more), this pattern is not adversarial to PQCache.

\textbf{\textsc{Cost of Discriminator required to find \texorpdfstring{$b$}{b}}}

Figure~\ref{fig:kvpress-dca-cost-gap} shows the cost (in \$) required to find adversarial data points $b$ for state-of-the-art models from the Gemini and GPT families. We use GEPA and Cursor as the evolution harnesses. Across models, the evolved solutions consistently increase the gap between unapproximated and approximated inference, suggesting that the discriminator recipe transfers across providers and model capabilities. Surprisingly, lighter models such as Gemini Flash Lite, Gemini Flash, and GPT Luna are substantially more effective at generating adversarial data than larger models such as GPT Sol and Gemini Pro. This difference is even more pronounced in cost: lighter models can discover adversarial data for a fraction of a dollar, while larger models often struggle under the same instructions.

\textbf{Cost Certificate.}
The cost required to break an algorithm can serve as a certificate of robustness. Approximate inference algorithms may deviate from unapproximated inference on some inputs, but the cost of discovering such failures provides a quantitative measure of how difficult they are to expose under a specified adversarial generation and evaluation procedure. We refer to this measure as the \emph{cost certificate}. For example, Figure~\ref{fig:kvpress-dca-cost-gap} shows that FastKVZip can be broken with GPT Luna (GEPA) for \$0.20 on average at a gap of 0.5, while DCA can be broken for less than \$0.01. The certificate also enables comparisons within a domain: at 2\% sparsity, PQCache can be broken for less than \$0.01, whereas breaking Oracle Topk requires \$5.06.

\textbf{\textsc{Closing the Loop}}

\looseness=-1 We run GAL with both a discriminator, implemented using GEPA evolution, and a generator, implemented using Cursor/Claude evolution. We run two complete GAL iterations on two problem domains: context extension, using DCA as the state-of-the-art baseline, and KV compression, using Compactor~\citep{chari2025compactor}. The results are presented in Figure~\ref{fig:gal_results}. In both domains, GAL discovers algorithms that not only improve performance on the adversarial data generated by the discriminator, but also improve over the standard benchmarks that constitute $B_0$.

\textbf{KV Compression.} We use Qwen3-4B-Instruct-2507 at $4\times$ compression. In round 1, the discriminator hid a ledger needle among many lookalike rows in the middle of the context, causing Compactor's global top-$k$ selection to discard the relevant ledger entry. The generator countered by introducing a stratified budget that allocates compression capacity across local chunks, ensuring that all parts of the context receive priority. In round 2, the discriminator constructed a multi-hop chain using sparse route IDs that the updated Gen-1 algorithm still dropped. The generator countered by combining KeyDiff~\citep{park2026keydiff}, which preserves distinctive keys, with attention protection and value-norm signals, allowing both hop IDs and semantic content to survive under a unified scoring function.

\textbf{Context Extension.} We use Llama3-8B-Instruct with $4\times$ context extension. In round 1, the discriminator constructed a needle-in-a-haystack task in which the target needle appeared in the middle of the context among lookalike distractors, exposing a failure mode of DCA described in Figure~\ref{fig:dca-box}. The generator countered by incorporating ideas from Self-Extend, a recent approach to context extension. In round 2, the discriminator filled the context with many near-identical records, making the task more challenging under DCA's position adjustments. The generator countered by introducing continuous positioning and preserving the original value scale to better fit the extended context within the model's trained window.

\input{tex_figures/combined_ladder}

%% file: tex_figures/box1.tex
\galpacktrue
\newlength{\galpackchrome}
\newlength{\galpackpanel}
\newlength{\galpackimg}
\setlength{\galpackchrome}{48pt}
\setlength{\galpackpanel}{\dimexpr(0.50\textheight-\galpackchrome)/4\relax}
\setlength{\galpackimg}{\dimexpr\galpackpanel-10pt\relax}
\renewcommand{\galpanelht}{\the\galpackpanel}
\renewcommand{\galimght}{\the\galpackimg}
\renewcommand{\galbarht}{1.1cm}

\begin{figure*}[t]
\centering
\input{utility/svg_box}\end{figure*}

\begin{figure*}[t]
\noindent
\input{utility/dca_box}\vspace{-15pt}
\end{figure*}

%% file: utility/svg_box.tex
\ifgalpack\else
\begin{figure*}[t]
\centering
\fi
\begin{tcolorbox}[
  enhanced,
  colback=boxgray,
  colframe=black!55,
  boxrule=0.55pt,
  arc=\ifgalpack 3pt\else 5pt\fi,
  width=\textwidth,
  left=\ifgalpack 2pt\else 4pt\fi,
  right=\ifgalpack 2pt\else 4pt\fi,
  top=\ifgalpack 1pt\else 3pt\fi,
  bottom=\ifgalpack 1pt\else 4pt\fi,
  title={\textbf{Figure~\ref{fig:svg-box} Sparse Attention via SVG}},
  fonttitle=\ifgalpack\small\else\normalsize\fi,
  coltitle=black,
  colbacktitle=black!8,
  attach boxed title to top left={xshift=8pt,yshift=-2mm},
  boxed title style={
    boxrule=0.45pt,
    colframe=black!55,
    arc=3pt,
    left=5pt,
    right=5pt,
    top=\ifgalpack 1pt\else 2pt\fi,
    bottom=\ifgalpack 1pt\else 2pt\fi
  }
]
\footnotesize
\newcommand{\svgpanel}[3]{%
\begin{tcolorbox}[
  colback=white,
  colframe=black!35,
  colbacktitle=#1,
  coltitle=black,
  title=\textbf{#2},
  fonttitle=\scriptsize,
  fontupper=\scriptsize,
  boxrule=0.35pt,
  arc=2pt,
  left=2pt,
  right=2pt,
  top=1pt,
  bottom=1pt,
  equal height group=svg-top-panels,
  valign=#3
]}

\begin{minipage}[t]{0.492\textwidth}
\svgpanel{orange!15}{1. Adversarial prompt, validation, and results}{top}
\raggedright
\ifgalpack
Extreme macro \textbf{slow-motion hovering hummingbird}
from liquid gallium\ldots\
\textbf{dense fractal metallic feathers},
\textbf{razor-thin highlights}; wings blur\ldots\
molten beads, frost, pinpoint reflections\ldots
\else
Extreme macro \textbf{slow-motion of a hovering hummingbird}
cast from liquid gallium\ldots\
Its \textbf{feathers are a dense fractal fan of hair-thin metallic barbs},
each edge catching a \textbf{razor-thin highlight};
its wings blur\ldots\
Beads of molten metal\ldots\
frost crystallizes\ldots\
thousands of pinpoint reflections\ldots 
\fi
\par\medskip
\hrule height 0.3pt
\smallskip
\textbf{Validation.}\enspace
The dense model produces a good video.
\par\smallskip
\hrule height 0.3pt
\smallskip
\textbf{Results.}\par
\centering
\begin{tabular}{@{}lccc@{}}
\toprule
 & PSNR $\uparrow$ & SSIM $\uparrow$ & LPIPS $\downarrow$ \\
\midrule
Score & $15.57$ & $0.60$ & $0.45$ \\
\bottomrule
\end{tabular}
\end{tcolorbox}
\end{minipage}\hspace{3pt}%
\begin{minipage}[t]{0.492\textwidth}
\svgpanel{blue!10}{2. Dense and sparse video snapshots}{top}
\centering
\begin{minipage}[t]{0.485\linewidth}
\centering
\textbf{Dense model}\par
\includegraphics[width=\linewidth,height=\galimght]{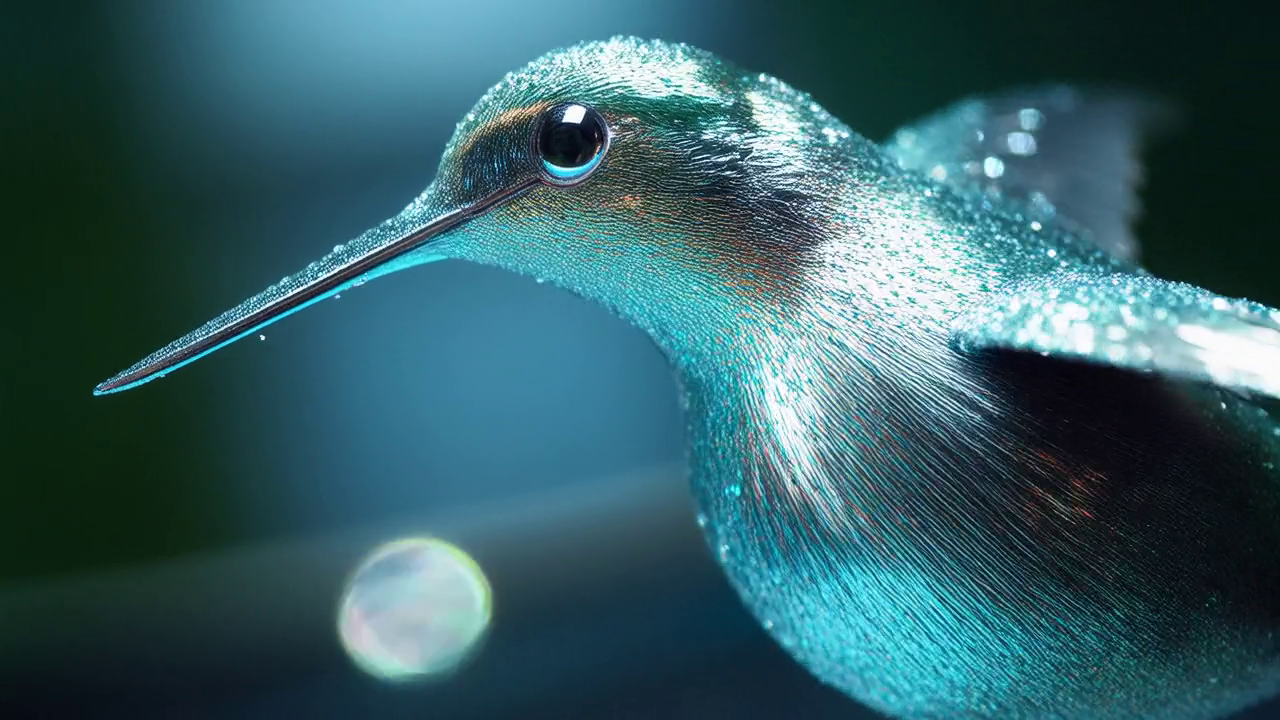}
\end{minipage}\hfill
\begin{minipage}[t]{0.485\linewidth}
\centering
\textbf{Sparse model}\par
\includegraphics[width=\linewidth,height=\galimght]{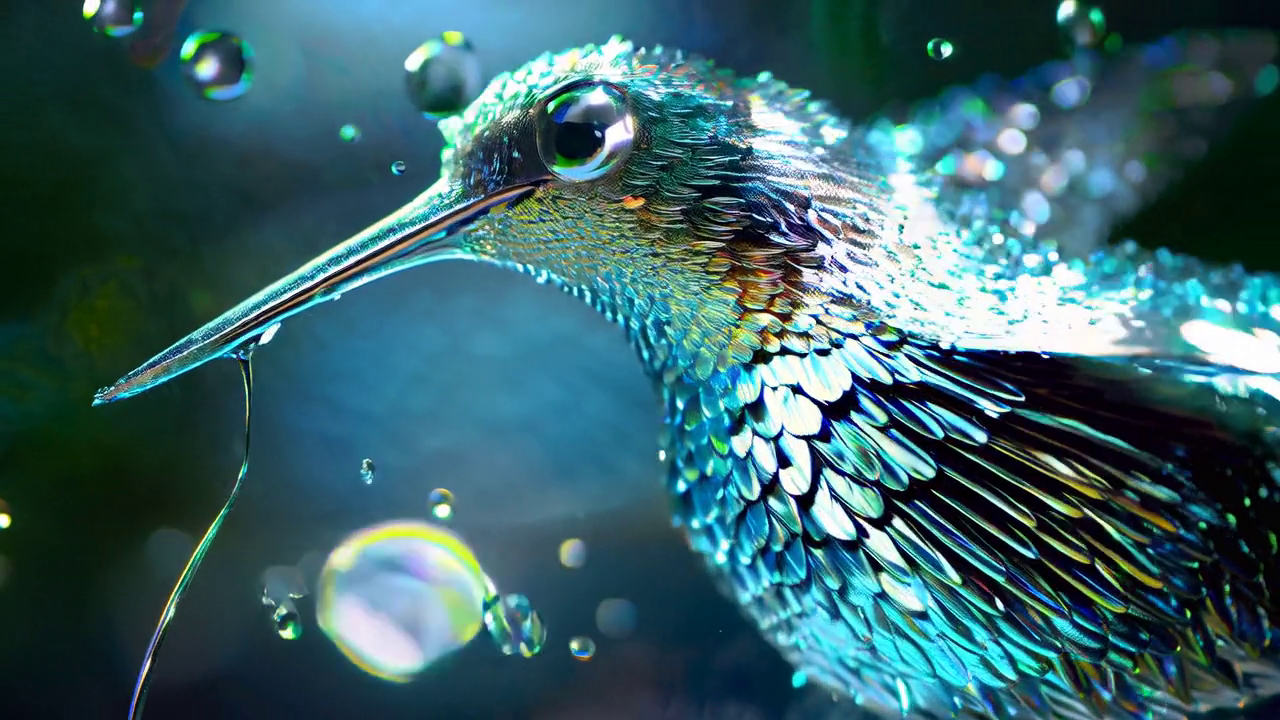}
\end{minipage}
\end{tcolorbox}
\end{minipage}

\par\vspace{\ifgalpack-2pt\else2pt\fi}
\begin{tcolorbox}[
  colback=yellow!12,
  colframe=black!35,
  boxrule=0.35pt,
  arc=2pt,
  left=3pt,
  right=3pt,
  top=2pt,
  bottom=2pt,
  fontupper=\scriptsize
]
\textbf{Explanation.}\enspace
The prompt (full prompt in appendix) is very detailed, requiring high-resolution video and computation and complex light and mirrored surface interaction.
\enspace\textbf{Settings.}\enspace
We use the Wan2.1-T2V-1.3B model for 81 frames at
$720\times1280$ pixels. The sparse version uses 30\% sparsity; all other
settings follow the SVG recommendations.
Dense and sparse models use the same seed and sampling schedule.
\end{tcolorbox}
\end{tcolorbox}
\ifgalpack
\vspace{-8pt}
\refstepcounter{figure}
\label{fig:svg-box}
\else
\refstepcounter{figure}
\label{fig:svg-box}
\end{figure*}
\fi

%% file: utility/dca_box.tex
\ifgalpack\else
\begin{figure*}[t]
\centering
\fi
\begin{tcolorbox}[
  enhanced,
  colback=boxgray,
  colframe=black!55,
  boxrule=0.55pt,
  arc=\ifgalpack 3pt\else 5pt\fi,
  width=\textwidth,
  left=\ifgalpack 2pt\else 4pt\fi,
  right=\ifgalpack 2pt\else 4pt\fi,
  top=\ifgalpack 1pt\else 3pt\fi,
  bottom=\ifgalpack 1pt\else 4pt\fi,
  title={\textbf{Figure~\ref{fig:dca-box} Context Extension via DCA}},
  fonttitle=\ifgalpack\small\else\normalsize\fi,
  coltitle=black,
  colbacktitle=black!8,
  attach boxed title to top left={xshift=8pt,yshift=-2mm},
  boxed title style={
    boxrule=0.45pt,
    colframe=black!55,
    arc=3pt,
    left=5pt,
    right=5pt,
    top=\ifgalpack 1pt\else 2pt\fi,
    bottom=\ifgalpack 1pt\else 2pt\fi
  }
]
\footnotesize
\newcommand{\dcapanel}[3]{%
\begin{tcolorbox}[
  colback=white,
  colframe=black!35,
  colbacktitle=#1,
  coltitle=black,
  title=\textbf{#2},
  fonttitle=\scriptsize,
  fontupper=\scriptsize,
  boxrule=0.35pt,
  arc=2pt,
  left=2pt,
  right=2pt,
  top=1pt,
  bottom=1pt,
  equal height group=dca-top-panels,
  valign=#3
]}

\begin{minipage}[t]{0.492\textwidth}
\dcapanel{orange!15}{1. Adversary, validation, and results}{top}
\begin{lstlisting}[style=dcapy,numbers=none,xleftmargin=0pt,frame=none,
  basicstyle=\ttfamily\scriptsize,aboveskip=0pt,belowskip=0pt,
  breaklines=true,breakindent=6pt,columns=fullflexible]
def build_adversary():
 s = make_goldstr(); on_chunk_boundary(s)
 add_distractors(); return instance
\end{lstlisting}
\par\smallskip
\hrule height 0.3pt
\smallskip
\textbf{Validation}: 1. \textbf{By construction:} create Q/A before placing
the target. 2. \textbf{Baseline control:} solved at native length $W$.
\par\smallskip
\hrule height 0.3pt
\smallskip
\textbf{Results.}\par
\centering
\begin{tabular}{@{}lccc@{}}
\toprule
 & $\mathcal{C}@N$ & $\mathcal{A}@\lambda=4\times N$ & Gap \\
\midrule
$B_0$ & $1.00$ & $0.90$ & $0.10$ \\
$B_1$ & $0.95$ & $0.50$ & $0.45$ \\
\bottomrule
\end{tabular}
\end{tcolorbox}
\end{minipage}\hspace{3pt}%
\begin{minipage}[t]{0.492\textwidth}
\dcapanel{blue!10}{2. DCA position mapping}{top}
\centering
\scalebox{1}[0.85]{%
  \includegraphics[width=0.75\linewidth]{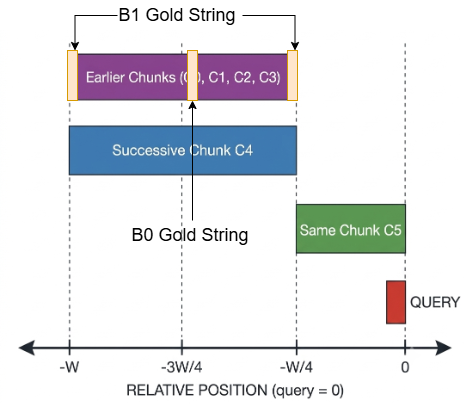}%
}
\end{tcolorbox}
\end{minipage}

\par\vspace{\ifgalpack-2pt\else2pt\fi}
\begin{tcolorbox}[
  colback=yellow!12,
  colframe=black!35,
  boxrule=0.35pt,
  arc=2pt,
  left=3pt,
  right=3pt,
  top=2pt,
  bottom=2pt,
  fontupper=\scriptsize
]
\textbf{Explanation.}\enspace
The evolved generator moves the target to a DCA chunk boundary and adds same-format distractors. Position remapping can fragment the target in the query's effective view, while locally coherent distractors remain competitive.
\enspace\textbf{Settings.}\enspace
We use Llama3-8B-Instruct with an 8K native context window and
$4\times$ context extension.
\end{tcolorbox}
\end{tcolorbox}
\ifgalpack
\vspace{-8pt}
\refstepcounter{figure}
\label{fig:dca-box}
\else
\refstepcounter{figure}
\label{fig:dca-box}
\end{figure*}
\fi

%% file: tex_figures/box2.tex
\begin{figure*}[t]
\noindent
\input{utility/kvpress_box}\par\vspace{-0.5pt}%
\input{utility/pqcache_box}\vspace{-15pt}
\end{figure*}

%% file: utility/kvpress_box.tex
\ifgalpack\else
\begin{figure*}[t]
\centering
\fi
\begin{tcolorbox}[
  enhanced,
  colback=boxgray,
  colframe=black!55,
  boxrule=0.55pt,
  arc=\ifgalpack 3pt\else 5pt\fi,
  width=\textwidth,
  left=\ifgalpack 2pt\else 4pt\fi,
  right=\ifgalpack 2pt\else 4pt\fi,
  top=\ifgalpack 1pt\else 3pt\fi,
  bottom=\ifgalpack 1pt\else 4pt\fi,
  title={\textbf{Figure~\ref{fig:kvpress-box} KV Compression via FastKVZip}},
  fonttitle=\ifgalpack\small\else\normalsize\fi,
  coltitle=black,
  colbacktitle=black!8,
  attach boxed title to top left={xshift=8pt,yshift=-2mm},
  boxed title style={
    boxrule=0.45pt,
    colframe=black!55,
    arc=3pt,
    left=5pt,
    right=5pt,
    top=\ifgalpack 1pt\else 2pt\fi,
    bottom=\ifgalpack 1pt\else 2pt\fi
  }
]
\footnotesize
\newcommand{\kvpanel}[3]{%
\begin{tcolorbox}[
  colback=white,
  colframe=black!35,
  colbacktitle=#1,
  coltitle=black,
  title=\textbf{#2},
  fonttitle=\scriptsize,
  fontupper=\scriptsize,
  boxrule=0.35pt,
  arc=2pt,
  left=2pt,
  right=2pt,
  top=1pt,
  bottom=1pt,
  equal height group=kv-top-panels,
  valign=#3
]}

\begin{minipage}[t]{0.492\textwidth}
\kvpanel{orange!15}{1. Adversary, validation, and results}{top}
\begin{lstlisting}[style=dcapy,numbers=none,xleftmargin=0pt,frame=none,
  basicstyle=\ttfamily\scriptsize,aboveskip=0pt,belowskip=0pt,
  breaklines=true,breakindent=6pt,columns=fullflexible]
def build_adversary():
  fresh, stale = make_ledgers()
  place(fresh, mid); place(stale, near_end)
  return instance
\end{lstlisting}
\par\smallskip
\hrule height 0.3pt
\smallskip
\textbf{Validation}: 1. \textbf{By construction:} a solution exists.
2. \textbf{Baseline control:} the model with full KV performs well.
\par\smallskip
\hrule height 0.3pt
\smallskip
\textbf{Results.}\par
\centering
\begin{tabular}{@{}lc@{}}
\toprule
 & Accuracy \\
\midrule
Uncompressed & $0.93$ \\
FastKVZip & $0.35$ \\
\midrule
Gap & $\mathbf{0.58}$ \\
\bottomrule
\end{tabular}
\end{tcolorbox}
\end{minipage}\hspace{3pt}%
\begin{minipage}[t]{0.492\textwidth}
\kvpanel{blue!10}{2. Standard and evolved benchmarks}{top}
\centering
\textbf{$B_0$: Standard benchmark}\par
\includegraphics[width=0.75\linewidth]{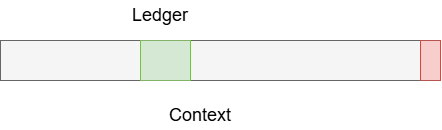}
\par\smallskip
\textbf{$B_1$: Stale ledger at an unadvantageous position}\par
\includegraphics[width=0.75\linewidth]{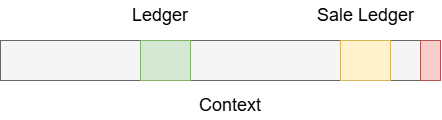}
\end{tcolorbox}
\end{minipage}

\par\vspace{\ifgalpack-2pt\else2pt\fi}
\begin{tcolorbox}[
  colback=yellow!12,
  colframe=black!35,
  boxrule=0.35pt,
  arc=2pt,
  left=3pt,
  right=3pt,
  top=2pt,
  bottom=2pt,
  fontupper=\scriptsize
]
\textbf{Explanation.}\enspace
Stale and fresh information is harder to distinguish during compression. To
make retrieval harder, the stale ledger is placed near the prompt and the
fresh ledger farther away.
\enspace\textbf{Settings.}\enspace
We use FastKVZip with Qwen3-8B, a 32K-token context, and
$4\times$ KV-cache compression.
\end{tcolorbox}
\end{tcolorbox}
\ifgalpack
\vspace{-8pt}
\refstepcounter{figure}
\label{fig:kvpress-box}
\else
\refstepcounter{figure}
\label{fig:kvpress-box}
\end{figure*}
\fi

%% file: utility/pqcache_box.tex
\ifgalpack\else
\begin{figure*}[t]
\centering
\fi
\begin{tcolorbox}[
  enhanced,
  colback=boxgray,
  colframe=black!55,
  boxrule=0.55pt,
  arc=\ifgalpack 3pt\else 5pt\fi,
  width=\textwidth,
  left=\ifgalpack 2pt\else 4pt\fi,
  right=\ifgalpack 2pt\else 4pt\fi,
  top=\ifgalpack 1pt\else 3pt\fi,
  bottom=\ifgalpack 1pt\else 4pt\fi,
  title={\textbf{Figure~\ref{fig:pqcache-box} Sparse Decode Attention via PQCache}},
  fonttitle=\ifgalpack\small\else\normalsize\fi,
  coltitle=black,
  colbacktitle=black!8,
  attach boxed title to top left={xshift=8pt,yshift=-2mm},
  boxed title style={
    boxrule=0.45pt,
    colframe=black!55,
    arc=3pt,
    left=5pt,
    right=5pt,
    top=\ifgalpack 1pt\else 2pt\fi,
    bottom=\ifgalpack 1pt\else 2pt\fi
  }
]
\footnotesize
\newcommand{\pqpanel}[3]{%
\begin{tcolorbox}[
  colback=white,
  colframe=black!35,
  colbacktitle=#1,
  coltitle=black,
  title=\textbf{#2},
  fonttitle=\scriptsize,
  fontupper=\scriptsize,
  boxrule=0.35pt,
  arc=2pt,
  left=2pt,
  right=2pt,
  top=1pt,
  bottom=1pt,
  equal height group=pq-top-panels,
  valign=#3
]}

\begin{minipage}[t]{0.492\textwidth}
\pqpanel{orange!15}{1. Adversary, validation, and results}{top}
\begin{lstlisting}[style=dcapy,numbers=none,xleftmargin=0pt,frame=none,
  basicstyle=\ttfamily\scriptsize,aboveskip=0pt,belowskip=0pt,
  breaklines=true,breakindent=6pt,columns=fullflexible]
def build_adversary():
  s = make_gold_str(); place(s, center)
  return instance
\end{lstlisting}
\par\smallskip
\hrule height 0.3pt
\smallskip
\textbf{Validation}: 1. \textbf{By construction:} a solution exists.
2. \textbf{Baseline control:} dense attention retrieves the gold string.
\par\smallskip
\hrule height 0.3pt
\smallskip
\textbf{Results.}\par
\centering
\begin{tabular}{@{}lc@{}}
\toprule
 & Accuracy \\
\midrule
Dense attention & $0.80$ \\
PQCache ($2\%$) & $0.00$ \\
\midrule
Gap & $\mathbf{0.80}$ \\
\bottomrule
\end{tabular}
\end{tcolorbox}
\end{minipage}\hspace{3pt}%
\begin{minipage}[t]{0.492\textwidth}
\pqpanel{blue!10}{2. Standard and evolved benchmarks}{top}
\centering
\textbf{$B_0$: Standard benchmark}\par
\includegraphics[width=0.75\linewidth]{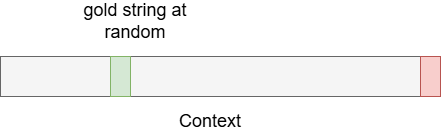}
\par\smallskip
\textbf{$B_1$: Evolved benchmark}\par
\includegraphics[width=0.75\linewidth]{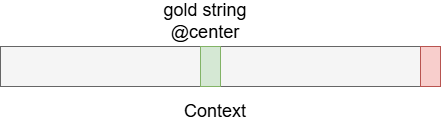}
\end{tcolorbox}
\end{minipage}

\par\vspace{\ifgalpack-2pt\else2pt\fi}
\begin{tcolorbox}[
  colback=yellow!12,
  colframe=black!35,
  boxrule=0.35pt,
  arc=2pt,
  left=3pt,
  right=3pt,
  top=2pt,
  bottom=2pt,
  fontupper=\scriptsize
]
\textbf{Explanation.}\enspace
Attention scores are typically high near the extremities and low in the
middle. Extensive training largely mitigates this lost-in-the-middle effect
for dense attention, but it resurfaces under top-$k$ sparse attention such as
PQCache.
\enspace\textbf{Settings.}\enspace
We use Qwen3-4B-Instruct-2507 with PQCache at 2\% sparsity.
\end{tcolorbox}
\end{tcolorbox}
\ifgalpack
\vspace{-8pt}
\refstepcounter{figure}
\label{fig:pqcache-box}
\else
\refstepcounter{figure}
\label{fig:pqcache-box}
\end{figure*}
\fi

%% file: utility/cost_plots_combined.tex
\begin{figure*}[t]
  \centering
  \includegraphics[width=\textwidth]{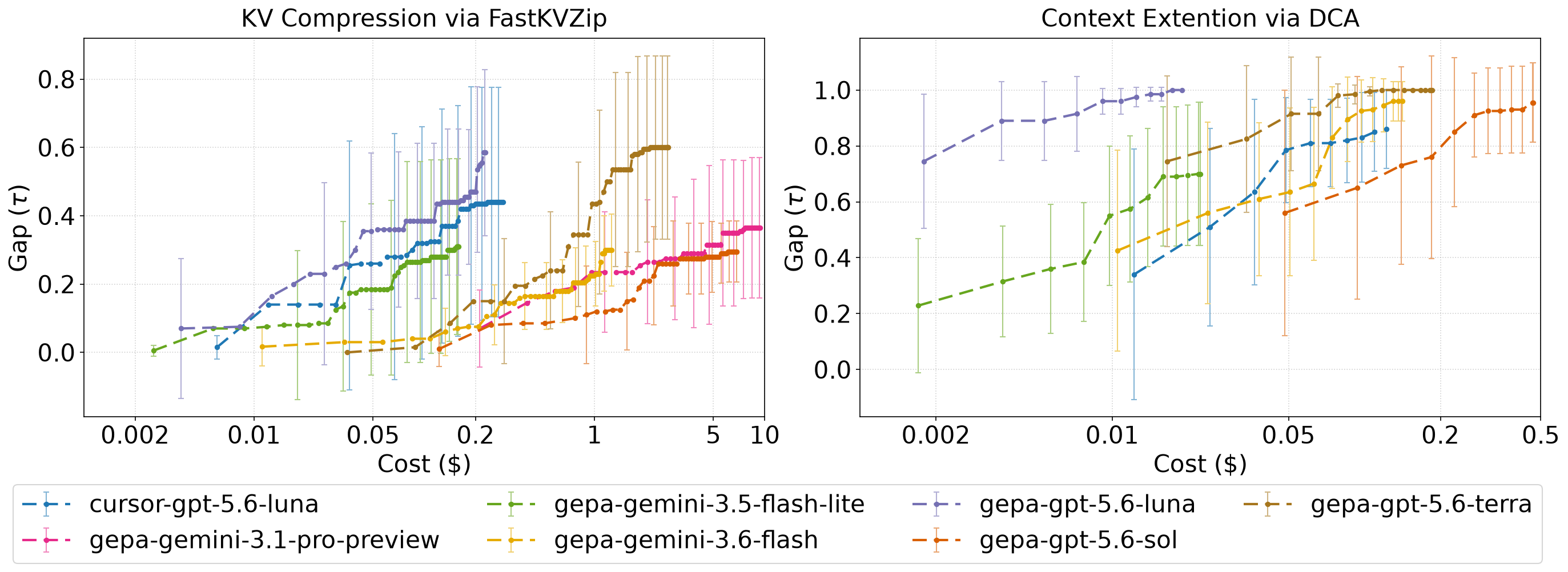}
  \caption{\footnotesize Best gap ($\tau$) versus cost for KV compression (FastKVZip, left) and context extension (DCA, right). Gap is averaged over 10 runs per setting.}
  \label{fig:kvpress-dca-cost-gap}
  
\vspace{-30pt}
\end{figure*}

\galpackfalse
\renewcommand{\galpanelht}{4.35cm}
\renewcommand{\galimght}{3.55cm}
\renewcommand{\galbarht}{1.5cm}

%% file: tex_figures/combined_ladder.tex
\begin{figure}[t]
\centering
\footnotesize

\begin{minipage}{\linewidth}
\centering
\setlength{\tabcolsep}{3pt}
\begin{tabular}{|l|cc|cccccc|c|}
\hline
 & \multicolumn{2}{c|}{\textbf{Evolved benchmarks}}
 & \multicolumn{6}{c|}{\textbf{Paper benchmarks (RULER-4K hard)}}
 & \textbf{Avg.} \\
\hline
\textbf{algorithm}
 & \textbf{$B_1$} & \textbf{$B_2$}
 & \textbf{qa\_1} & \textbf{qa\_2} & \textbf{vt} & \textbf{cwe} & \textbf{fwe} & \textbf{niah\_mv}
 & \textbf{RULER Avg.} \\
\hline
\textbf{no-press} & 1.00 & 0.95 & 0.84 & 0.60 & 1.00 & 0.962 & 0.8867 & 0.995 & 0.8806 \\
\hline
\textbf{Compactor} & \textcolor{red}{0.25} & \textcolor{red}{0.45} & 0.54 & 0.48 & 1.00 & 0.76 & 0.80 & 0.92 & 0.7500 \\
\hline
\textbf{Gen-1} & \textcolor{ForestGreen}{0.85} & \textcolor{red}{0.25} & 0.62 & 0.54 & 1.00 & 0.736 & 0.7933 & 0.92 & 0.7682 \\
\hline
\textbf{Gen-2} & \textcolor{ForestGreen}{0.95} & \textcolor{ForestGreen}{1.00} & 0.54 & 0.48 & 1.00 & 0.728 & 0.8133 & 0.985 & 0.7577 \\
\hline
\end{tabular}

\vspace{1pt}
\end{minipage}

\vspace{8pt}

\begin{minipage}{\linewidth}
\centering
\setlength{\tabcolsep}{3pt}
\begin{tabular}{|l|ccc|cc|cc|cc|c|c|}
\hline
 & \multicolumn{3}{c|}{\textbf{Evolved benchmarks}}
 & \multicolumn{8}{c|}{\textbf{Paper benchmarks}} \\
\hline
 & & &
 & \multicolumn{2}{c|}{\textbf{passkey}}
 & \multicolumn{2}{c|}{\textbf{needle}}
 & \multicolumn{2}{c|}{\textbf{PG19 perplexity}}
 & \textbf{Coursera}
 & \textbf{ScienceFiction} \\
\hline
\textbf{algorithm}
 & \textbf{seed} & \textbf{$B_1$} & \textbf{$B_2$}
 & \textbf{16k} & \textbf{32k}
 & \textbf{16k} & \textbf{32k}
 & \textbf{16k} & \textbf{32k}
 & {\scriptsize$\sim$9.9k} & {\scriptsize$\sim$15.2k} \\
\hline
\textbf{Full}   & 0.00 & 0.00 & 0.00 & 0.00 & 0.00 & 0.00 & 0.00 & 134.07 & 1025.96 & 40.0 & 50.0 \\
\hline
\textbf{DCA}    & 0.90 & \textcolor{red}{0.00} & \textcolor{red}{0.40} & 1.00 & 1.00 & 1.00 & 1.00 & 12.51 & 12.45 & 58.0 & 68.0 \\
\hline
\textbf{Gen-1}  & 1.00 & \textcolor{ForestGreen}{1.00} & \textcolor{red}{0.57} & 1.00 & 1.00 & 1.00 & 1.00 & 12.39 & 12.19 & 60.0 & 70.0 \\
\hline
\textbf{Gen-2}  & 1.00 & \textcolor{ForestGreen}{0.97} & \textcolor{ForestGreen}{0.83} & 1.00 & 1.00 & 1.00 & 1.00 & 12.36 & 12.12 & 58.0 & 74.0 \\
\hline
\end{tabular}
\end{minipage}
\vspace{-6pt}
\caption{\footnotesize \textbf{GAL on KV Compression (top) and Context Extension(Bottom)}
Compactor Settings:  $4\times$ compression, model= Qwen3-4B-Instruct-2507 using Cursor evolution harness. DCA Settings: $4\times$ extension with model=Llama3 with 8K context. The generated algorithms in both cases perform well on adversarial benchmarks while maintaining, and in some cases improving, performance on standard benchmarks.}
\vspace{-8pt}
\label{fig:gal_results}

\end{figure}

%% file: sections/appendix.tex
\input{sections/gepa_setup}
\input{sections/cursor_generator_setup}




%% file: sections/gepa_setup.tex
\section{Gepa Setup}
\label{sec:gepa_setup}

\subsection{Purpose and Scope}
\label{app:gepa-purpose}

This appendix describes how \textsc{Adversearch-Agent} uses GEPA~\cite{agrawal2026gepa} as the outer optimization loop for discovering adversarial candidates. GEPA iteratively mutates a \emph{candidate artifact} under a fixed, harness-defined evaluation protocol. Depending on the task, this artifact is either (i)~a \emph{code candidate}, a function that constructs adversarial inputs or evaluation examples, or (ii)~a \emph{prompt candidate}, a single adversarial prompt string. In both cases, GEPA evolves the candidate text, while the harness is responsible for materializing and evaluating the resulting artifact.

Each candidate is scored by a task-specific evaluator that scores both the baseline system and an experimental configuration (for example, full-precision versus quantized inference) and returns a scalar objective score together with a reflection summary. The role of GEPA is to search the space of candidate programs so as to maximize this objective within a fixed metric-call budget.

\subsection{Integration Architecture}
\label{app:gepa-architecture}

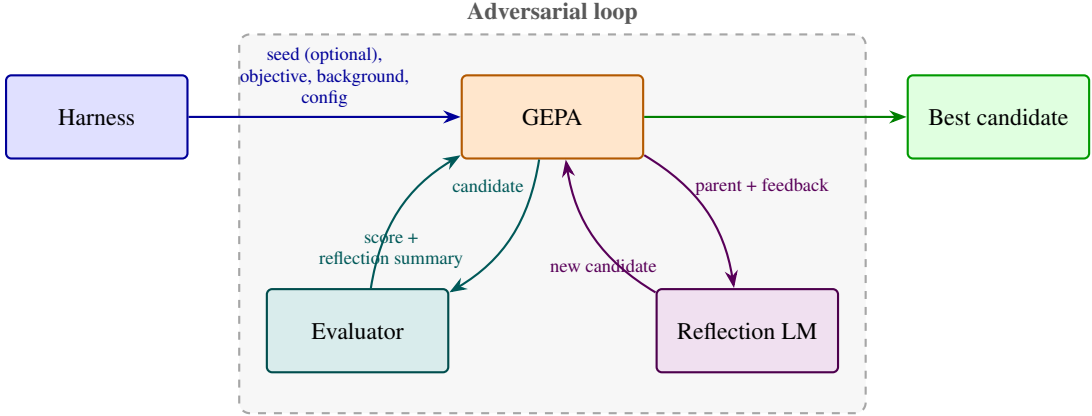
\begin{figure}[h!]
\centering
\begin{tikzpicture}[
  box/.style={draw, rounded corners=2pt, align=center, minimum height=1.1cm, minimum width=2.4cm, font=\small, thick},
  harness/.style={box, fill=blue!12, draw=blue!60!black},
  gepa/.style={box, fill=orange!20, draw=orange!70!black},
  eval/.style={box, fill=teal!15, draw=teal!60!black},
  refl/.style={box, fill=violet!12, draw=violet!60!black},
  outbox/.style={box, fill=green!12, draw=green!60!black},
  arrow/.style={-{Stealth}, thick},
  lbl/.style={font=\scriptsize, align=center}
]
  \node[harness] (agent) {Harness};
  \node[outbox, right=9.5cm of agent] (out) {Best candidate};

  \node[gepa, right=3.6cm of agent] (gepa) {GEPA};
  \node[eval, below left=1.7cm and 0.15cm of gepa] (eval) {Evaluator};
  \node[refl, below right=1.7cm and 0.15cm of gepa] (refl) {Reflection LM};

  \begin{scope}[on background layer]
    \node[
      draw=gray!70,
      dashed,
      thick,
      rounded corners=4pt,
      fill=gray!6,
      fit=(gepa)(eval)(refl),
      inner xsep=10pt,
      inner ysep=15pt,
      label={[font=\small\bfseries, text=gray!70!black]above:Adversarial loop}
    ] (loop) {};
  \end{scope}

  \draw[arrow, blue!60!black]
    (agent) -- node[above, lbl] {seed (optional),\\objective, background,\\config} (gepa);

  \draw[arrow, teal!70!black]
    (gepa) to[bend left=25] node[left, lbl, pos=0.15] {candidate} (eval);
  \draw[arrow, teal!70!black]
    (eval) to[bend left=25] node[below, lbl, pos=0.4] {score +\\reflection summary} (gepa);

  \draw[arrow, violet!70!black]
    (gepa) to[bend left=25] node[right, lbl, pos=0.3] {parent + feedback} (refl);
  \draw[arrow, violet!70!black]
    (refl) to[bend left=25] node[below, lbl, pos=0.4] {new candidate} (gepa);

  \draw[arrow, green!50!black] (gepa) -- (out);
\end{tikzpicture}
\caption{The harness provides seed(optional), objective, background, run configuration, and a harness evaluator to GEPA. In each iteration of the adversarial loop, GEPA evaluates the latest candidate and builds reflection summary, it then uses a reflection LLM to propose the next candidate by mutating over a chosen parent candidate.}
\label{fig:gepa-adversearch}
\end{figure}

Figure~\ref{fig:gepa-adversearch} shows the overall GEPA integration in Adversearch-Agent. A harness supplies the fixed task setup---candidate contract, objective, background, evaluator, optional seed, and GEPA configuration---and GEPA is invoked through \texttt{optimize\_anything} (package version~\texttt{0.1.1}). Inside the adversarial loop, GEPA repeatedly selects a parent candidate, evaluates it to obtain a score and reflection summary, and asks a reflection LLM to propose a mutated child. This cycle continues until the metric-call budget is exhausted, after which GEPA returns the best candidate.

\paragraph{Parent selection and candidate acceptance}

Each evaluation returns a harness-defined scalar \emph{aggregate score} that measures how close a candidate is to meeting the objective. GEPA accepts a child into the parent pool only when this score improves on its parent; rejected proposals are discarded.

GEPA selects the next parent from the pool using Pareto-based selection over the multi-objective score vector returned by the evaluator. The axes of this frontier are harness-defined. For example, in the KVPress harness the evaluator uses  \texttt{baseline\_accuracy}, and \texttt{gap} ( baseline\_accuracy -- experiment\_accuracy) so the Pareto frontier trades off keeping baseline accuracy high while increasing the accuracy gap under KV compression; both axes matter for parent selection.

The scalar aggregate score is still used to decide if a new candidate gets accepted, while the Pareto axes determine which parents are eligible to generate the next reflective mutation.

\subsection{Practical Learnings for Using Adversearch}
\label{app:gepa-learnings}
These notes summarize practical design lessons from configuring Adversearch for GEPA-based adversarial search.

\begin{itemize}
\item \textbf{No solution hints:}
The task background should not provide hints at attack strategies that can bias the exploration.

\item \textbf{Constraints and validation:}
Background and objective should state goals and hard candidate constraints. The reflection LM may still ignore constraints, so additional harness-side validation and rejection remain essential.

\item \textbf{Seeds bias the search:}
The initial candidate strongly shapes later proposals. We added an exploration encouraging prompt addon and also ran seedless optimizations to reduce hand-seed bias. In seedless mode, the reflection LM itself is used to generate the seed candidate based on the background and objective.

\item \textbf{Pareto axes representing objective:}
Extra frontiers can enlarge the parent pool, but contradictory higher-is-better axes can make search oscillate. For example, adding \texttt{(1-experiment)} rewarded candidates that also broke the baseline; axes should instead favor selectively breaking the experiment while preserving baseline behavior.

\item \textbf{Diversified seeds, then independent evolution:}
In preliminary multi-island runs we generated diverse seeds and evolved each independently under a shared budget, aiming for several refined candidates rather than one collapsed lineage.

\item \textbf{Post-hoc diversity metrics:}
We also experimented with preliminary diversity measures over candidates and
IDEAs, including LLM judges and embedding cosine similarity, used for
analysis rather than closed-loop control.
\end{itemize}

%% file: sections/cursor_generator_setup.tex
\section{Generator Cursor Agent Setup}
\label{sec:cursor_generator_setup}

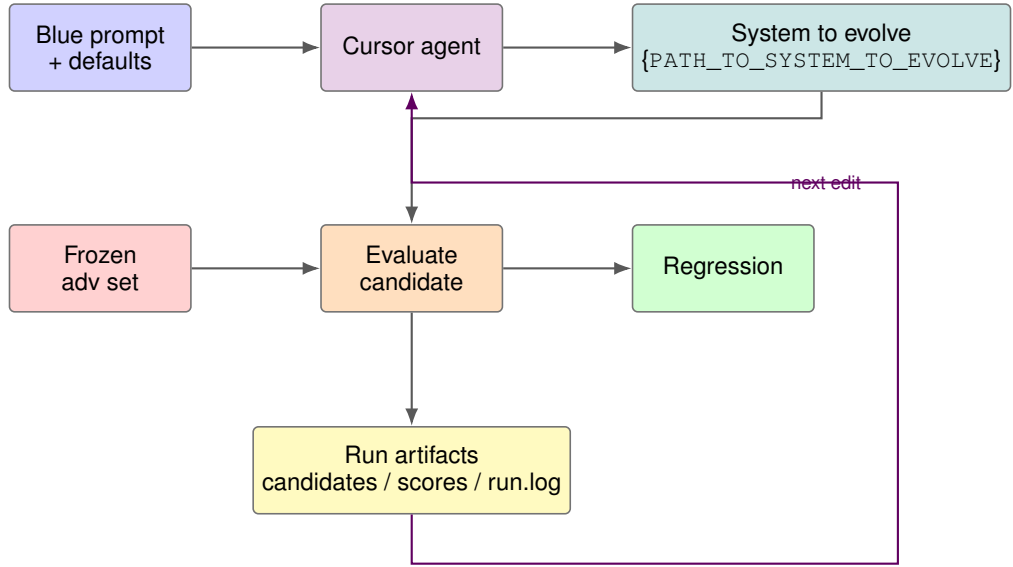
\begin{figure}[t]
\centering
\begin{tikzpicture}[
  >=Latex,
  box/.style={
    draw=black!55, rounded corners=2pt, align=center,
    minimum height=1.15cm, minimum width=2.4cm,
    font=\small\sffamily, line width=0.6pt
  },
  arr/.style={->, thick, draw=black!65},
  feedback/.style={->, thick, draw=violet!75!black},
  node distance=1.5cm and 1.7cm
]
  \node[box, fill=blue!18]    (prompt) {Blue prompt\\+ defaults};
  \node[box, fill=violet!18, right=of prompt] (agent) {Cursor agent};
  \node[box, fill=teal!20,   right=of agent]  (algo)
    {System to evolve\\\{\texttt{PATH\_TO\_SYSTEM\_TO\_EVOLVE}\}};

  \node[box, fill=orange!25, below=1.75cm of agent] (eval) {Evaluate\\candidate};
  \node[box, fill=red!18,    left=of eval]          (adv)  {Frozen\\adv set};
  \node[box, fill=green!18,  right=of eval]         (reg)  {Regression};

  \node[box, fill=yellow!30, below=1.5cm of eval]   (art)
    {Run artifacts\\candidates / scores / run.log};

  \draw[arr] (prompt) -- (agent);
  \draw[arr] (agent)  -- (algo);
  \draw[arr] (algo.south) -- ++(0,-0.35) -| (eval.north);
  \draw[arr] (adv)  -- (eval);
  \draw[arr] (eval) -- (reg);
  \draw[arr] (eval) -- (art);

  \draw[feedback]
    (art.south) -- ++(0,-0.65)
    -| ([xshift=1.1cm]reg.east)
    -- ($(reg.east |- eval.north) + (1.1cm,0.55cm)$)
    -| (agent.south)
    node[pos=0.12, right, font=\scriptsize\sffamily, text=violet!75!black]
      {next edit};
\end{tikzpicture}
\caption{Cursor Blue Team loop: edit the system, evaluate on a frozen adversarial set
and regression suite, then iterate from logged scores.}
\end{figure}

\subsection{Agent prompt}

\lstset{
  basicstyle=\ttfamily\small,
  breaklines=true,
  frame=single,
  backgroundcolor=\color{gray!8},
  columns=fullflexible,
  keepspaces=true,
}
\begin{lstlisting}[escapeinside=||]
## Task
Improve `{PATH_TO_SYSTEM_TO_EVOLVE}` so accuracy on the primary adversarial set rises.
Held-out regression floors in `{DEFAULTS}` are a **gated** check: run them when a
candidate looks promising, not on every iteration (evals are slow).

Evolve the algorithm by editing files under `{PATH_TO_SYSTEM_TO_EVOLVE}/`. |$\ldots$|
Evaluate every candidate on the **primary adv set** via `{EVAL_SCRIPT}` |$\ldots$|

**Default loop:** primary-only. Optimize for higher `adv_accuracy` /
`objective_score` on that set vs `target_adv` in `{DEFAULTS}`.

**When to run held-out regression (your call):** when the primary result meets or
approaches `target_adv`, is a new best by a clear margin, or you need a regression
sanity check before committing to a direction. |$\ldots$|

Stop only when iteration budget reaches `MAX_CANDIDATES`. Don't early exit if
the goal is met or if progress stalls. Don't generate all candidates at once |---|
later edits must be informed by scores so far.

## Boundaries
- Stay inside this machine's project trees needed for Blue Team only.
- You may read and write:
  - `{PATH_TO_SYSTEM_TO_EVOLVE}/`
  - `{RUN_DIR}/`
  - |$\ldots$|
- Do not modify the frozen adversarial set.
- Do not read the Red Team tree |$\ldots$|
- Do not cite, summarize, imitate, or otherwise use prior Blue runs for this task.
- Write run artifacts only under `{RUN_DIR}/`.
- Do not use evolution frameworks (GEPA / OpenEvolve / AlphaEvolve) for this loop.

## Evaluation command
### 1) Materialize adv once, then freeze
### 2) Iterate: edit system |$\rightarrow$| evaluate
### 3) Gate: evaluate on regression suite

|$\ldots$|
\end{lstlisting}